\documentclass[journal,transmag]{IEEEtran}

\usepackage{hyperref}
\usepackage{multirow}
\usepackage{amsmath}
\usepackage{graphicx}
\usepackage{pgffor}
\usepackage{setspace}
\usepackage{xcolor}

\newcommand{\etal}{\textit{et al.}}

\begin{document}

%\begin{frontmatter}

%\title{Analyses and Enhancement of Fairness in Face Presentation Attack Detection}
\title{Fairness in Face Presentation Attack Detection}

\author{Meiling Fang$^{1,2,3}$, Wufei Yang$^{3}$, Arjan Kuijper$^{1,3}$, Vitomir \u{S}truc$^{4}$, Naser Damer$^{1,3}$ \\
$^{1}$Fraunhofer Institute for Computer Graphics Research, Darmstadt, Germany\\
$^{2}$College of Information Engineering, Yangzhou University, Yangzhou, China\\
$^{3}$Department of Computer Science, TU Darmstadt,
Darmstadt, Germany\\
$^{4}$ Faculty of Electrical Engineering, University of Ljubljana, Ljubljana, Slovenia \\
Email: meiling.fang@yzu.edu.cn
}

\maketitle

\begin{abstract}
Face recognition (FR) algorithms have been proven to exhibit discriminatory behaviors against certain demographic and non-demographic groups, raising ethical and legal concerns regarding their deployment in real-world scenarios. Despite the growing number of fairness studies in FR, the fairness of face presentation attack detection (PAD) has been overlooked, mainly due to the lack of appropriately annotated data. To avoid and mitigate the potential negative impact of such behavior, it is essential to assess the fairness in face PAD and develop fair PAD models. To enable fairness analysis in face PAD, we present a Combined Attribute Annotated PAD Dataset (CAAD-PAD), offering seven human-annotated attribute labels. 
Then, we comprehensively analyze the fairness of PAD and its relation to the nature of the training data and the Operational Decision Threshold Assignment (ODTA) through a set of face PAD solutions. 
Additionally, we propose a novel metric, the Accuracy Balanced Fairness (ABF), that jointly represents both the PAD fairness and the absolute PAD performance. The experimental results pointed out that female and faces with occluding features (e.g. eyeglasses, beard, etc.) are relatively less protected than male and non-occlusion groups by all PAD solutions. To alleviate this observed unfairness, we propose a plug-and-play data augmentation method, FairSWAP, to disrupt the identity/semantic information and encourage models to mine the attack clues. The extensive experimental results indicate that FairSWAP leads to better-performing and fairer face PADs in 10 out of 12 investigated cases. CAAD-PAD is publicly available \footnote{\url{https://github.com/meilfang/FairnessFacePAD}}.
\end{abstract}


\IEEEpeerreviewmaketitle

%\linenumbers

\section{Introduction}
\label{sec:intro}
Face recognition (FR) \cite{arcface,DBLP:journals/pr/BoutrosDKK22} {has made significant advances} in recent years and {has been successfully} deployed in various practical scenarios, such as unlocking mobile phones or automated border control. 
{However, FR is vulnerable to presentation attacks (PA) \cite{DBLP:conf/eccv/ZhangYLYYSL20}. An attacker can use a PA, such as a printed photo\cite{casia-fasd}, a replayed video \cite{oulu}, 3D mask \cite{padisi-Face}, or masked spoof faces \cite{DBLP:journals/pr/FangDKK22,DBLP:conf/fgr/FangBKD21}}, to impersonate someone or obfuscate his/her own identity. Therefore, face presentation attack detection (PAD) is crucial to mitigate a major vulnerability of FR algorithms. 
{Over the last few decades, machine learning and deep learning techniques has been successfully applied in various computer vision fields, such as image recognition \cite{DBLP:journals/pami/YuTZRT22}, object detection \cite{DBLP:journals/pami/ChenZKLZQSJ23}, pose estimation \cite{DBLP:journals/tip/HongYWTW15,DBLP:journals/tii/HongYZJL19}, as well as PAD \cite{DBLP:journals/pr/WangYZ23, DBLP:conf/icb/PurnapatraSBDYM21}. However, these deep learning-based methods heavily rely on the scale and diversity of training data, and such data-driven approaches are known to exhibit unfairness towards certain data categories \cite{nist_biasAI_2022}.} 
%Many works {have} achieved remarkable progress in face PAD performance by utilizing deep learning techniques that rely on the scale and diversity of training data. 
%Data-driven approaches, including FR algorithms, are known to be unfair between certain demographic and non-demographic groups \cite{DBLP:conf/aaai/0001MMV22,DBLP:journals/cviu/BeveridgeGPD09,DBLP:conf/icb/TerhorstKDKK20}.
Learnable models are strongly impacted by data-induced biases because they tend to optimize objectives toward the majority group, i.e. data represented with more samples in training datasets.
This commonly results in less optimized performance for minority groups, leading to unfair decisions. % \cite{}. 
{The assessment and enhancement of fairness} in biometric systems have gained increasing attention from {both} the research community and the general public \cite{9975333}. For example, many studies have investigated fairness in FR \cite{de2020fairness,Philipp_bias} and face quality \cite{DBLP:conf/icb/TerhorstKDKK20,DBLP:conf/cvpr/SVKAB19}.

Most studies on biometric fairness have concentrated {primarily} on demographic covariants, especially gender and race. Only a few studies \cite{Philipp_bias,DBLP:conf/icb/TerhorstKDKK20} investigated the impact of other variations, such as appearance traits, on biometric fairness. 
Moreover, the fairness of PAD has not been investigated, except for very limited studies on {iris PAD \cite{DBLP:conf/eusipco/FangDKK20} and face PAD \cite{alshareef2021study}}. {Fang \etal} \cite{DBLP:conf/eusipco/FangDKK20} addressed the gender fairness in iris PADs and the experimental results pointed out that female users were significantly less protected by the PAD in comparison to males.
{Alshareef \etal} \cite{alshareef2021study} considered gender fairness in face PAD by using ResNet50 \cite{resnet50} and VGG16 \cite{vgg16} on a limited PAD data. Both studies focused only on the gender fairness assessment and with very limited evaluation data, both in terms of size and diversity. The main reason is that the majority of publicly available PAD datasets do not contain information regarding demographic and non-demographic attributes, making it impossible to assess the fairness, let alone enhance potential unfairnesses. 
Furthermore, the fairness of the PADs in both studies \cite{DBLP:conf/eusipco/FangDKK20,alshareef2021study} {was} measured by {using} differential performance and outcome of PAD, i.e. no fairness metrics were applied. 
To date, only two very recent definitions of fairness in FR were proposed, Inequality Rate (IR) \cite{nist_IR_metric} and Fairness Discrepancy Rate (FDR) \cite{de2020fairness}.
IR \cite{nist_IR_metric} takes {the} ratio differences between {the} minimum and {the} maximum FR performance per group. However, IR has two drawbacks: 1) IR has no theoretical upper bound due to its multiplicative nature and exponential weights, 2) IR might be incomputable due to its ratio property, i.e. when {the} minimum FR performance (denominator) for any group is zero.
FDR \cite{de2020fairness} considers the maximum difference FR performance between any two groups based on a decision threshold calculated {across} all groups. However, FDR does not take absolute performance into account and thus might consider a fair but low-performing PAD "better" than an unfair PAD that performs close to perfect. 
%In our work, we adapt FDR metric to represent the fairness in PAD performance. 

% contribution:
To address these under-explored gaps in analyzing, representing, and enhancing fairness in face PAD, we present in this work the following contributions:
\begin{itemize}
\item We present a Combined Attribute Annotated PAD Dataset (CAAD-PAD) by combining six publicly available PAD datasets that include highly diverse PAs. To {facilitate} fairness assessment, {we provide} seven human-annotated attribute labels {that cover} both demographic and non-demographic attributes.
\item We assess the fairness of face PADs from two aspects: the nature of the training data and the Operational Decision Threshold Assignment (ODTA) based on data of different groups. 
\item We adapt the fairness metric, FDR, to face PAD and propose a novel metric, Accuracy Balanced Fairness (ABF), {that jointly represents both the fairness and the absolute PAD performance.}
\item To enhance the fairness of PADs, we propose a straightforward data augmentation solution, named FairSWAP. FairSWAP disrupts the identity and semantic information, boosting {the performance and fairness of PADs} in most cases, as will be demonstrated {through a comprehensive and diverse set of experiments.}
\end{itemize}

The rest of this work is organized as follows: Section \ref{sec:related_work} presents the essential background information regarding
fairness in automated decision systems and PAD. Section \ref{sec:caad_pad} introduces our CAAD-PAD, including descriptions of each used PAD dataset, human-annotation criteria, and experimental protocols designed to enable fairness analyses. Section \ref{sec:fairness_assessment} focuses on setups for fairness assessment and describes used PAD algorithms, evaluation metrics including our proposed ABF, and implementation details. Section \ref{sec:bias_mitigation} introduces the proposed FairSWAP for fairness enhancement and the corresponding implementation details. Section \ref{sec:fairness_assessment_baselines} and \ref{sec:bias_mitigation_PS} discuss the results of fairness assessment and enhancement of face PADs, respectively. A summary is presented in Section \ref{sec:conclusion}.

\section{Related work} % ready for Naser
\label{sec:related_work}

\begin{figure*}[htb]
\centering
\includegraphics[width=0.99\linewidth]{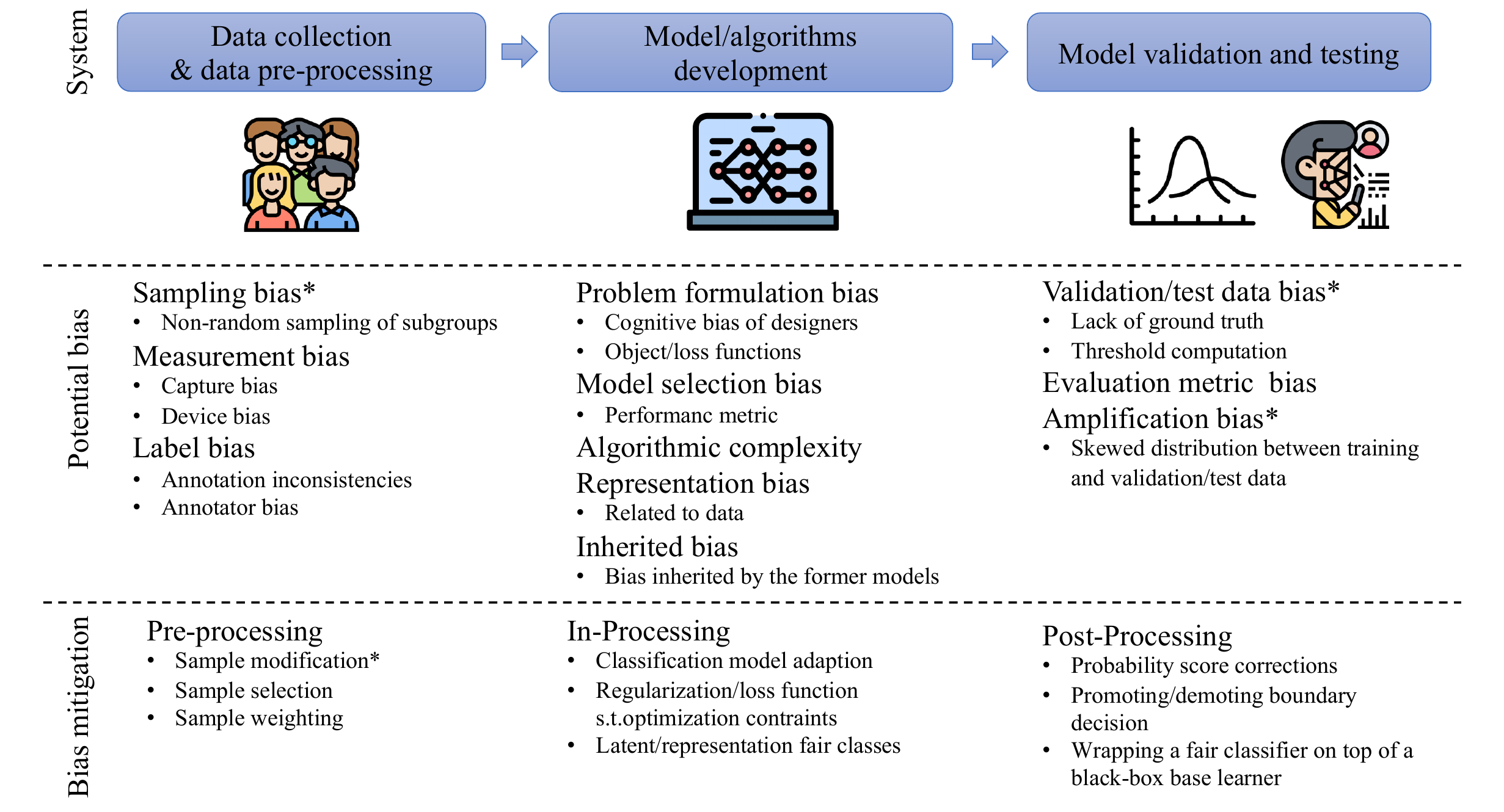}
\caption{Aspects effecting fairness in automated decision systems. $*$ indicates the addressed aspects in this work, where sampling bias, validation/test data bias, and amplification bias are relevant to the nature of training data, the ODTA on different data groups, and the training and test on group-disjoint data, respectively.}
\label{fig:overview_bias}
\end{figure*}

\subsection{Fairness in automated decision systems}
\label{ssec:background_fairness_biometric}
Biometric systems, as automated decision-making systems, have been widely deployed in recent decades. 
Building an automated decision-making system involves the following steps: data collection and pre-processing, model development, and model validation and testing, as illustrated in Figure \ref{fig:overview_bias}. 
{Biometric data}, such as face and iris traits, are collected by sensors and pre-processed {to serve} as input for learnable models. Model/algorithm development aims to summarize the pattern of biometric traits via supervised/unsupervised learning. The developed model is then evaluated with respect to certain metrics, such as accuracy and equal error rate (EER), among others.
As suggested in \cite{nist_biasAI_2022,de2020fairness}, {potential triggers of unfairness} may exist at every stage of an automated system, as listed in Figure \ref{fig:overview_bias} (* refers to aspects addressed in this work). In the data collection phase, {unfairness may arise due to sampling bias, resulting in} data imbalance, measurement bias related to the capture environments and sensors, and label bias attributed to annotators. 
Similarly, model/algorithm development could be human biased (e.g., problem formulation or objective function definition by human designer) or statistically biased (e.g., performance metric-based model selection or bias inherited from the training data or pre-processing), as argued in \cite{nist_biasAI_2022,DBLP:conf/kdd/Corbett-DaviesP17,danks2017algorithmic}.
Lastly, fairness issues could arise in model validation and testing given an account of potential validation/test bias, evaluation metric bias, or amplification bias \cite{nist_biasAI_2022}. Amplification bias occurs when the validation/test data is skewed compared to the training data distribution, which motivates our experimental protocol design in Section \ref{ssec:experimental_protocols}. In automated biometric systems, a threshold commonly computed from the validation (development) set is necessary to make a final decision. The unfair distribution of validation data could be transmitted to the threshold computation, which is an issue analyzed in this work.
Schwartz \etal \cite{nist_biasAI_2022} suggested that the dataset, evaluation, and human factor are the three critical challenges in machine learning fairness. Fairness issues caused by human factors are much more complex and multi-faceted, including societal and historical aspects. Therefore, this work focuses on studying the dataset- and ODTA-related fairness concerns and proposes a fairness enhancement solution {at} the data pre-processing level. 

%Biases in the validation data could transmitted to the threshold computation, as verified in our experiments. 
%The fairness issue in automated decision systems is complex and multi-faceted. 

\subsection{Fairness in face presentation attack detection}
\label{ssec:background_fairness_facePAD}
Recently, Drozdowski \etal \cite{drozdowski2020demographic} presented a comprehensive summary of the existing literature on fairness assessment and enhancement of biometric systems. This survey found that a majority of existing studies conducted fairness experiments using face traits and concentrated only on recognition algorithms, while the fairness of PADs was barely investigated.
{To date, the fairness of face PAD systems has received insufficient attention.} To the best of our knowledge, only one study \cite{DBLP:conf/eusipco/FangDKK20} explored the gender fairness of iris PAD systems and one work \cite{alshareef2021study} considered the gender fairness of face PAD by using ResNet50 \cite{resnet50} and VGG16 \cite{vgg16} on {a signle} face PAD dataset.
%, named Spoof in the Wild (SiW) \cite{DBLP:conf/cvpr/LiuJ018}. 
The fairness of demographic attributes (e.g., gender, age, race) and soft-biometric attributes (e.g., beards, hair, makeup, accessories) {in} face PAD systems {has} been extremely understudied. The main possible reason is the insufficient face PAD data and the lack of such attribute labels. Most existing PAD datasets, such as OULU-NPU \cite{oulu}, CASIA-MFSD \cite{casia-fasd}, MSU-MFS \cite{msu}, Idiap Replay-Attack \cite{replay}, were collected in laboratories {with} insufficient subjects (details can be found in Table \ref{tab: datasets-summariz}) and thus leading to limited variations in soft-biometric characteristics. A large-scale face PAD dataset, CelebA-spoof \cite{DBLP:conf/eccv/ZhangYLYYSL20}, provided some additional soft-biometric labels. However, after a comprehensive analysis on the data distribution, we found that data in attribute groups in CelebA-sooof is extremely unbalanced over bona fide and attack samples, making it impossible to design proper experimental protocols for PAD fairness analyses. For example, subjects with eyeglasses, makeup, or bangs are all bona fide samples. {This issue also exists} in the PADISI-Face \cite{padisi-Face} dataset. To address this issue, this work combined six publicly available face PAD datasets, consisting of print, replay, 3D mask and wax figure attacks, and provided seven human-annotated attribute labels to enable PAD fairness studies.

In addition to the lack of appropriate labeled data, there are no standard criteria {for assessing} the fairness of developed systems. Fang \etal \cite{DBLP:conf/eusipco/FangDKK20} studied demographic bias by adapting and reporting the differential performance and differential outcome as suggested for verification performance in \cite{DBLP:conf/btas/HowardSV19}. Differential performance, as in \cite{DBLP:conf/eusipco/FangDKK20}, measures the difference in the bona fide or attack decision distribution between specific attribute groups independently of any decision threshold, while differential outcome describes the difference in APCER or BPCER rates between different demographic groups relative to a group-specific decision threshold. Alshareef \etal \cite{alshareef2021study} measured the demographic bias and fairness of PAD solutions by observing the difference in PAD performance, such as Area Under the Receiver operating characteristic curve (ROC-AUC), accuracy, EER, and APCER/BPCER. Both studies assessed fairness either using group-specific decision thresholds or being independent of any decision threshold. 
However, using such group-specific thresholds is not fair for different groups, as mentioned in \cite{de2020fairness}, and measuring fairness independently from thresholds is not realistic. Such problems were raised and explored in FRs in \cite{Krishnapriya2022}.

Our work also assesses the fairness associated with ODTA of face PADs by exploring the PAD performance under different attribute group decision thresholds (corresponds to the marked points of the model validation/testing phases in Figure \ref{fig:overview_bias}). To bridge {the gaps in} the fairness measurement, Pereira and Marcel \cite{de2020fairness} introduced FDR to evaluate and compare  fairness between biometric verification systems. Unlike measuring the fairness by reporting differential performance \cite{DBLP:conf/eusipco/FangDKK20,alshareef2021study,drozdowski2020demographic}, FDR assessed the trade-off between model performances by assuming a single "fair" decision threshold for all demographic groups. 
Therefore, in this study, FDR is adapted to assess the fairness of face PADs (details in Section \ref{ssec:evaluation_metrics}). However, FDR does not consider absolute performance, thus a completely fair but low-performing PAD will be considered "better" than a slightly unfair PAD that performs perfectly, according to FDR. 
{As a result, we propose the ABF metric to further link the PAD performance and fairness.}

\section{Combined Attribute Annotated PAD Dataset}
\label{sec:caad_pad}
% incuding datasets, annotation, data distribution and protocols

Extensive research \cite{de2020fairness,Philipp_bias,drozdowski2020demographic,DBLP:journals/cviu/BeveridgeGPD09} has shown that recognition systems exhibit bias, i.e. subjects in a certain demographic or non-demographic groups are more accurately recognized than others. However, most of these research efforts focus {only on exploring} \cite{de2020fairness,Philipp_bias,DBLP:journals/cviu/BeveridgeGPD09} or {mitigating} \cite{DBLP:conf/cvpr/Gong0021,DBLP:conf/iwbf/TerhorstTDKK20,DBLP:conf/wacv/BruverisMGM20} bias in FR algorithms, {with no} detailed investigation of demographic and non-demographic bias in face PAD systems. 
Besides the contemporary nature of biometric bias studies, {one} possible reason for this is the lack of sufficient PAD data with soft-biometric labels. {To address this issue, we combine} six face PAD datasets and provide publicly released corresponding annotations covering demographic and non-demographic attributes, named Combined Attribute Annotated PAD Dataset (CAAD-PAD).
{We present a} detailed description of each selected PAD dataset, the criteria of annotations, the distribution of CAAD-PAD, and the experimental protocols for fairness assessment in the following Section \ref{ssec:datasets}, \ref{ssec:annotation_criteria}, and \ref{ssec:experimental_protocols}, respectively.

\subsection{Datasets} % done 1st
\label{ssec:datasets}

%=====================================================================
\begin{table}[htb]
\centering
\resizebox{0.49\textwidth}{!}{
\begin{tabular}{c|c|c|c|c|c}
\hline 
Dataset & Year & \# BF/attack  & \# Sub & Attack types & Attribute label \\ \hline \hline
CASIA-FASD \cite{casia-fasd} & 2012 & 150/450 (V) & 50 & 1 Print, 1 Replay  & No \\ \hline
Replay-Attack \cite{replay} & 2012 & 200/1,000 (V) & 50  & 1 Print, 2 Replay & No  \\ \hline
MSU-MFSD \cite{msu} & 2015 & 70/210 (V) & 35 & 1 Print, 2 Replay & No \\ \hline
HKBU-MARs \cite{hkbu} & 2016 & 120/60 (V) & 12  & 2 3D masks  & No \\ \hline
OULU-NPU \cite{oulu}  & 2017 & 1,980/3,960 (V) & 55 & 2 Print,2 Replay  & No \\ \hline
SWFFD \cite{swffd} & 2019 & 2,300/2,300 (I) & 745 & 1 3D face & No \\ \hline
CelebA-Spoof \cite{DBLP:conf/eccv/ZhangYLYYSL20} & 2020 & \begin{tabular}[c]{@{}c@{}}184,407/\\ 377,168(I)\end{tabular}       & 10,177  & \begin{tabular}[c]{@{}c@{}}3 Print, 3 Replay, \\ 1 3D, 3 Paper Cut\end{tabular}  & Yes (40) \\ \hline
PADISI-Face \cite{padisi-Face}  & 2021 & 1,105/924 (V) & 360 & \begin{tabular}[c]{@{}c@{}}1 Print, 4 Mask, \\1 Makeup, 1 Tattoo, \\2 Partial\end{tabular}  & Yes (4) \\ \hline
CAAD-PAD (our) & 2022 & \begin{tabular}[c]{@{}c@{}} 2,510/5,680(V)\\ 2,300/2,300(I)\end{tabular} & 947 & \begin{tabular}[c]{@{}c@{}}3 Print, 2 Replay, \\ 2 3D masks, 1 wax face\end{tabular} & Yes (7) \\ \hline
\end{tabular}}
\caption{Summary of the main existing face PAD datasets. "V" and "I" refers to video and image sample, respectively. "BF" indicates the bona fide samples. The number following the "Yes" is the number of attribute label types.
CAAD-PAD (our) combines CASIA-FASD \cite{casia-fasd}, Replay-Attack \cite{replay}, MSU-MFSD \cite{msu}, HKBU-MARs \cite{hkbu}, OULU-NPU \cite{oulu} and SWFFD \cite{swffd}.  Despite the rich annotations of CelebA-Spoof \cite{DBLP:conf/eccv/ZhangYLYYSL20}, the provided attributes are extremely unbalanced over bona fide and attack, making it impossible to design proper protocols for fairness assessment. The PADISI-Face \cite{padisi-Face} has the same issue and is of a much smaller scale. These issues are discussed in details in Section \ref{sec:caad_pad} and motivates the need for our CAAD-PAD dataset. }
\label{tab: datasets-summariz}
\end{table}
%=====================================================================

\begin{figure*}[htbp]
\centering
\includegraphics[width=0.8\textwidth]{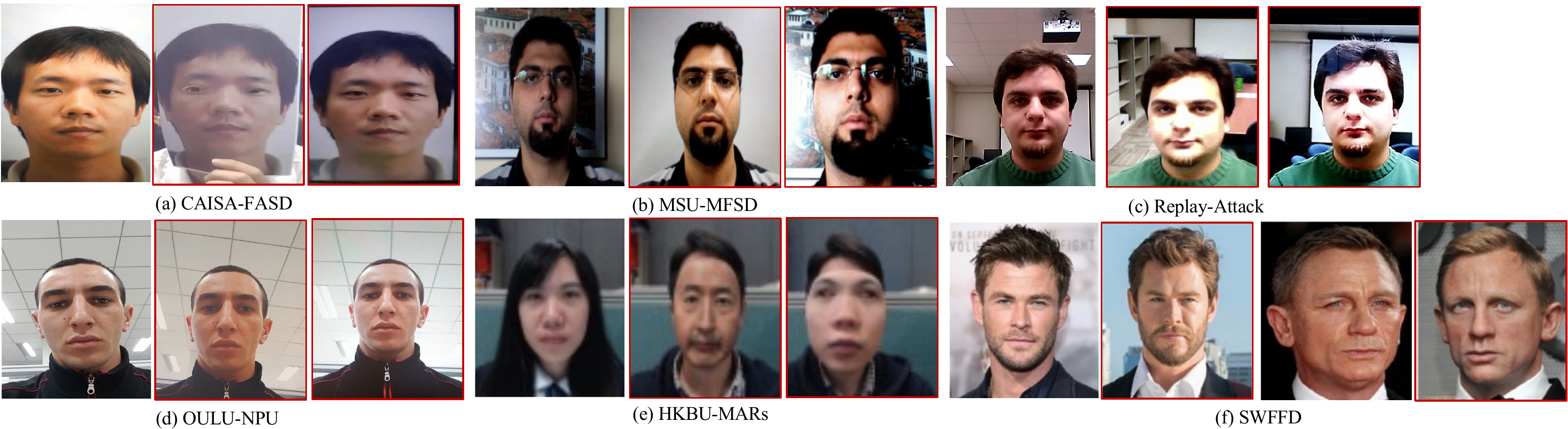}
\caption{Samples from the datasets used to built our CAAD-PAD (attacks in red frame).}
\label{fig:dataset_examples}
\end{figure*}

The selected six face PAD datasets (to build our CAAD-PAD) are presented in detail in this section. 
Table \ref{tab: datasets-summariz} summarizes the information of these PAD datasets, {along with} two conventional PAD datasets that contain attribute labels. 
CelebA-Spoof dataset \cite{DBLP:conf/eccv/ZhangYLYYSL20} involving rich annotations (as listed in Table \ref{tab: datasets-summariz}) is considered unsuitable for the goal of this work due to the extreme imbalance of its data, making it impossible to design proper experimental protocols for fairness study.
For example, subjects with attributes, such as eyeglasses, makeup and bangs, are all bona fide samples in the CelebA-Spoof dataset. A similar issue exists in the PADISI-Face \cite{padisi-Face} dataset. 
To facilitate fairness assessment of face PAD, we selected the following publicly available face PAD datasets (samples in Figure \ref{fig:dataset_examples}): CASIA-FASD \cite{casia-fasd}, Replay-Attack \cite{replay}, MSU-MFSD \cite{msu}, HKBU-MARs \cite{hkbu}, OULU-NPU \cite{oulu} and SWFFD \cite{swffd}, to be composed into our CAAD-PAD. 

This combination of datasets is chosen {for the following reasons}: 1) They consist of diverse PAs, including print and replay attacks, 3D mask attacks, and wax figure face attacks. 2) CASIA-FASD \cite{casia-fasd}, Replay-Attack \cite{replay}, MSU-MFSD \cite{msu}, and OULU-NPU \cite{oulu} are widely used in PAD studies \cite{DBLP:conf/wacv/FangDKK22,DBLP:conf/cvpr/LiuJ018,DBLP:conf/bmvc/DamerD16} which is the main problem in PAD.
3) HKBU-MARs \cite{hkbu} and SWFFD \cite{swffd} target the realistic mask attack problem, which is one of the practical PA problems in real-world applications.
The descriptions of each selected dataset are provided in the following:

\noindent \textit{CASIA-FASD} dataset \cite{casia-fasd} covers diverse potential attack types in the real world, including warped photo, cut photo, and video attacks. CASIA-FASD includes 600 videos across 50 subjects, where each subject has 12 videos (three bona fides and nine attacks). The videos consist of low, normal and high quality, where videos of low and normal quality have a resolution of $640 \times 480$, and high-quality videos have a resolution of $1280 \times 720$ pixels.  

\noindent \textit{Idiap Replay-Attack} \cite{replay} dataset contains 1,200 videos (200 bona fide and 1,000 attack) {from} 50 subjects. The videos were acquired in two sessions: controlled and adverse, {with no-uniform} background and illumination conditions. The attack videos were captured under two modes: {1) hand-based attack, where the attack device was held by a human hand, and 2) fixed-support attack, where the attack device was placed on fixed support.}

\noindent \textit{MSU-MFSD} \cite{msu} dataset consists of 440 videos (110 bona fide and 330 attacks) across two attack types, video replay and printed photo, of 55 subjects.

\noindent \textit{OULU-NPU} \cite{oulu} dataset was captured under realistic mobile authentication scenarios. {It comprises 5,940 videos from 55 subjects, with 1,980 bona fides and 3,960 attack videos.} The videos were captured using six different mobile phones under various illumination and backgrounds.

\noindent \textit{HKBU-MARs} \cite{hkbu} is a 3D mask PAD dataset {that includes masks generated} by two different techniques. First is the ThatsMyFace mask, which leverages 3D reconstruction and printing techniques to generate facial mask. The advantage of ThatsMyFace Mask is that the mask can be easily generated with a single customized image, but the skin texture is not well restored due to defects of the 3D printing technique. The second technique is the REAL-F mask \footnote{\url{https://real-f.jp}}, which provides a high-quality appearance and {closely resemble bona fide faces, as it can reconstruct} the skin texture, blood vessels of the eyes, and iris. This database contains 180 videos (120 bona fides and 60 attacks) of 12 subjects. An example of a bona fide sample, a ThatsMyFace attack, and a REAL-F mask are shown in Figure \ref{fig:dataset_examples} (e).

\noindent \textit{Single Wax Figure Face Database (SWFFD)} \cite{swffd} provides {realistic 3D PAs in the form of wax figure faces, complementing} the above face PAD datasets that cover replay, print, and other 3D attacks. {The SWFFD dataset, which features a high diversity of subjects sourced from the web, includes a total of 4,600 images, with 2,300 bona fide and 2,300 wax figure attack samples.} Examples of image of SWFFD \cite{swffd} are shown in Figure \ref{fig:dataset_examples} (f).

However, {the above selected PAD datasets only offer PAD labels without face attribute labels, limiting their ability to assess fairness in demographic and non-demographic groups. To facilitate fairness assessment,} we provide seven soft-biometric annotations in the following section.

\subsection{Face attribute annotation criteria} % done 1st
\label{ssec:annotation_criteria}

\begin{figure*}[thb]
\centering
\includegraphics[width=0.8\linewidth]{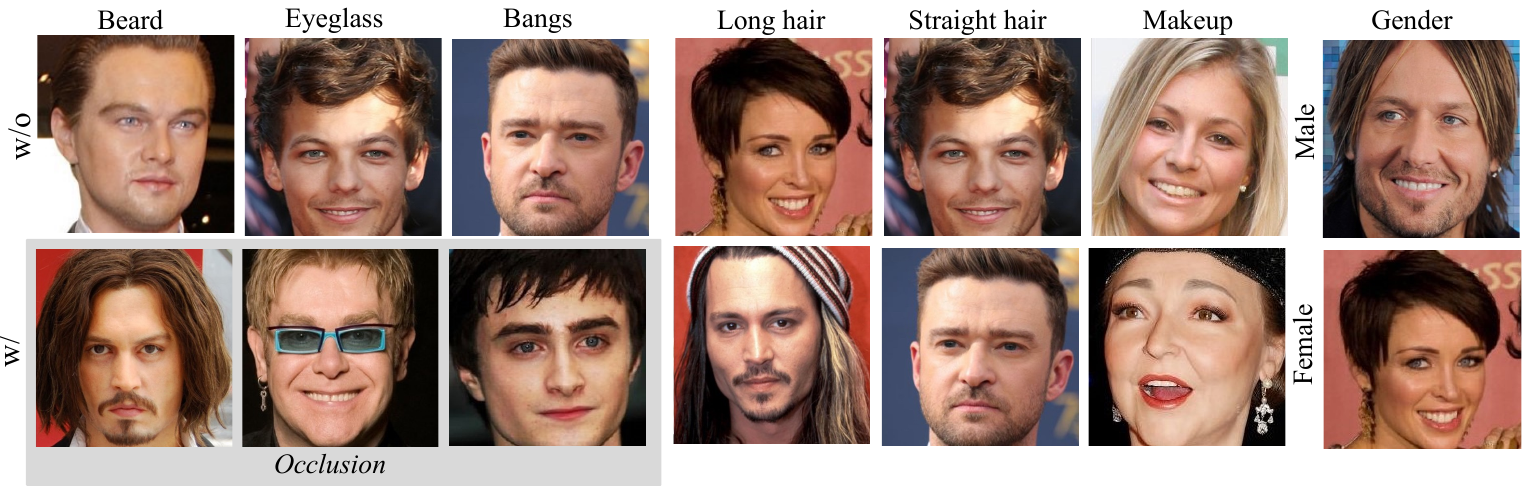}
\caption{Samples of each considered attribute. Samples are labeled as occlusion if they have beard, eyeglass, bangs, or a combination of them. Samples without beard, bangs, and not wearing eye glass are labeled as non-occlusion. The other attributes are individually labeled.}
\label{fig: attributes}
\end{figure*}

We manually annotate the CAAD-PAD dataset {based on} the following criterion (illustrated in Figure \ref{fig: attributes}): 1) \textbf{Gender}: Gender in our case is determined based on human perception of the face image. In this work, a subject is {assigned to either the} male or female group based on the majority decision of five annotators. These decisions might have also been influenced by previous knowledge of the subjects as some datasets included celebrities.
2) \textbf{Beard}: A subject with no visible hair coverage around the mouth or shaved with only light hair roots {is} labeled as having no beard. 3) \textbf{Eyeglasses}: A subject wearing eyeglasses in the face area is annotated as wearing eyeglasses, regardless of the type, shape, or color of the eyeglasses. Note that when the eyeglasses are in the area above the forehead, the sample is labeled as no eyeglasses because eyeglasses in such position will be removed after face detection and cropping. 4) \textbf{Bangs}: A subject with hair covering more than 15\% of the forehead is considered to have bangs. 5) \textbf{Makeup}: A subject with noticeable lipstick and eye shadow is categorized as wearing makeup. 6) \textbf{Long/short hair}: A subject with hair beyond the shoulder is considered to be in the long hair group. 7)  \textbf{Curly/Straight hair}: A subject with noticeable waves of hair is categorized into a curly hair group. Like gender, all annotations are based on the majority decision of five annotators.

Overall, the combined CAAD-PAD contains 8,190 videos and 4,600 images across 947 subjects, covering the seven binary attributes.

\begin{figure*}[thbp]
\centering
\includegraphics[width=0.9\textwidth]{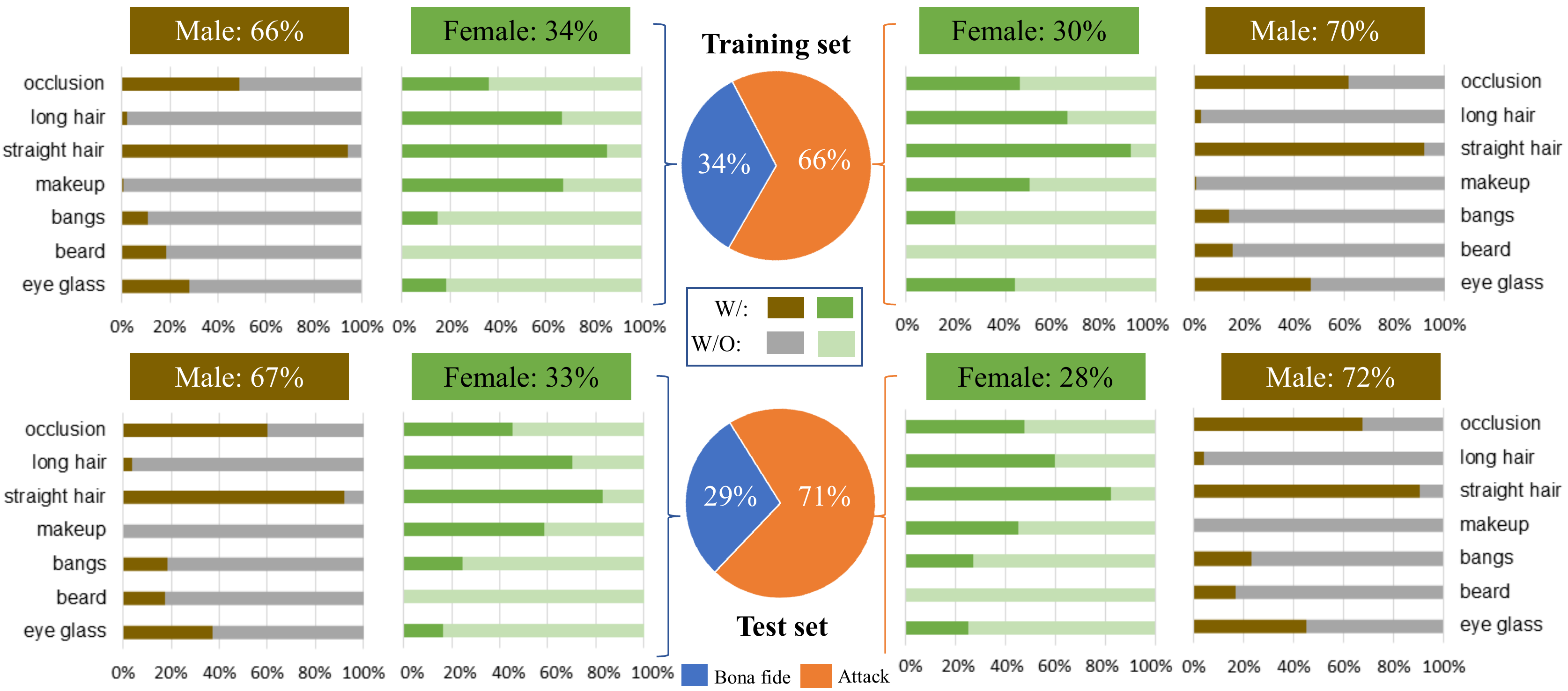}
\caption{Data distribution of the training and test set of CAAD-PAD. Most of the attributes are understandably not well-distributed over genders (but well distributed over bona fide and attacks), e.g. makeup of male sample, beard of female sample, long hair of male sample. Samples with occlusion and without occlusion are well-distributed over genders.}
\label{fig:data_distribution}
\end{figure*}

\subsection{Experimental protocols} % done 1st
\label{ssec:experimental_protocols}
\noindent \textbf{{Data distribution of CAAD-PAD:}}
We first present an in-depth analysis of the data distribution in the CAAD-PAD dataset (as shown in Figure \ref{fig:data_distribution}). 
The training set is {a combination of the} training and development subsets from {the} six selected datasets. Similarly, {the CAAD-PAD test set comprises the test subsets} from each of the selected datasets. Therefore, the subjects in the training and test set are disjoint. Further details are provided below:

\textit{Training set:}
{The training and development subsets of most selected datasets have pre-defined and identity-disjoint partitions, which we utilized in constructing the training set of our CAAD-PAD.} However, HKBU-MARs \cite{hkbu} does not provide such pre-split subsets. Therefore, we manually {construct identity-disjoint training and test subsets of HKBU-MARs} by considering subject attribute annotations with a subject ratio of 8:2 (train to test). {The training set of the CAAD-PAD contains a ratio of 1:1.9 bona fide and attack samples and a ratio of 1:2.1 female and male samples.} Samples with occlusion and without occlusion are well-distributed over genders, with a ratio close to 1:1. 
Although samples of some facial attributes (e.g., makeup, beard, long hair) are not well-distributed over genders, such samples are very well balanced between bona fide and attack in CAAD-PAD, unlike CelebA-Spoof \cite{DBLP:conf/eccv/ZhangYLYYSL20} and PADISI-Face \cite{padisi-Face} datasets.

\textit{Test set:} The test set {of CAAD-PAD} consists of data samples from the pre-defined test subsets in each dataset, {except HKBU-MARs \cite{hkbu}, which is }identity-disjoint split by us. {The distribution of the test set is similar to that of the training set,} as shown in Figure \ref{fig:data_distribution}.

\noindent \textbf{{Experimental Protocols:}}
{Once the CAAD-PAD dataset is constructed, we design the following protocols to} assess the PAD fairness on different groups. {These protocols aim to address} three aspects: gender, occlusion, and facial attributes. The impact of the nature of training data distribution on fairness is also examined. The designed protocols are as follows: 

\textit{Protocol-1} targets the fairness of PAD performance across gender groups. It consists of three sub-protocols: 1) Protocol 1.1: PAD solutions are trained on the entire training set of CAAD-PAD (including both female and male) and tested separately on female and male test sets. 2) Protocol 1.2: This protocol studies gender fairness when male data is unseen during training. Thus, PAD models are trained solely on female data in the training set and tested separately on the female and male data in the test set. 3) Protocol 1.3: Contrary to Protocol 1.2, PAD models in this protocol are trained only on males in the training set, and then tested separately on {the} female and male samples in the test set to study a gender fairness case where {the} female data is unseen during training.
    
\textit{Protocol-2} aims to assess the fairness of PAD performance across occlusion groups (occlusion and non-occlusion). Protocol-2 also consists of three sub-protocols: 1) Protocol 2.1: PAD models are trained on the entire training set (including samples with and without occlusion) and tested separately on occluded and non-occluded samples. 2) Protocol 2.2: This protocol explores fairness when the model only learns from occluded samples. Therefore, the PAD model is trained on samples with occlusion and tested separately on samples with and without occlusion. 3) Protocol 2.3: Conversely, the PAD solution is trained only on non-occluded samples in the training set and tested separately on samples with and without occlusion in the test set.
    
\textit{Protocol-3} explores the fairness of PAD performance across the remaining attribute groups (with/without eyeglass, beard, bangs, long/short hair, straight/short hair), respectively. As some attributes are unbalanced over genders, such as makeup and beard, we only train PAD models on the entire training set {and test them} on each attribute group separately, unlike the extended protocols over gender and occlusion.

% till here

\section{Fairness assessment} % ready for Naser
\label{sec:fairness_assessment}
This section first introduces the four PAD solutions {that serve as the bases of} our fairness assessment. Then, the metrics for PAD performance and fairness evaluation are presented. In addition, we adapt the fairness metric used and introduce a novel metric that links PAD biometric fairness with the worst PAD cases across all groups, as will be motivated in detail. Lastly, {we provide} implementation details to ensure reproducibility.

\subsection{PAD algorithms} % done 1st
\label{ssec:pad_algorithms}
To assess the fairness of PAD solutions, we adopt four diverse and well-established PAD solutions ranging from the texture-feature-based to deep-learning-based methods: LBP-MLP \cite{freitas2012lbp}, ResNet50 \cite{resnet50}, DeepPixBis \cite{deeppixbis}, LMFD \cite{DBLP:conf/wacv/FangDKK22}.

\noindent \textbf{LBP-MLP \cite{freitas2012lbp}:}
Considering that LBP is a widely used hand-crafted feature in earlier PAD studies \cite{DBLP:conf/icb/MaattaHP11,DBLP:conf/icip/BoulkenafetKH15,freitas2012lbp}, we use LBP features to investigate the fairness of face PAD. Following the highly influential work \cite{DBLP:conf/icb/MaattaHP11,DBLP:conf/icip/BoulkenafetKH15}, two LBP feature vectors are extracted {separately} from each image in RGB and HSV color space. These two feature vectors are then concatenated into one vector of dimension $60 \times 1$. To detect whether an input image is a bona fide or attack, the concatenated feature vector is fed to a simple Multi-Layer Perceptron (MLP) classifier consisting of only two fully-connected layers.

\noindent  \textbf{ResNet50 \cite{resnet50}:}
{The} residual learning framework is firstly proposed in \cite{resnet50} {to simplify and stabilize} the training of networks, {as the} computation complexity increases when the network grows. Considering that ResNet50 \cite{resnet50} {has been used} as a backbone architecture in many PAD methods \cite{DBLP:journals/tbbis/YuLSXZ21, DBLP:conf/eccv/ZhangYLYYSL20, DBLP:journals/tcsv/JiaLHGX21} and {has shown} good PAD performance, we use it in our experiments to further assess its fairness. 

\noindent  \textbf{DeepPixBis \cite{deeppixbis}:}
DeepPixBis \cite{deeppixbis} is the first work to adopt pixel-wise binary supervision to enhance PAD performance. The pixel-wise binary label on the output maps forces the network to learn fine-grained representation from different pixels/patches, {resulting in better performance. Given its success under the intra-dataset scenario, we include it in our PAD fairness assessment.}

\noindent  \textbf{LMFD-PAD \cite{DBLP:conf/wacv/FangDKK22}:}
LMFD-PAD \cite{DBLP:conf/wacv/FangDKK22} presents a dual-stream PAD framework, in which one stream learns features in the frequency domain while the other learns features in spatial color space. The benefit of this architecture is that features in the frequency domain {are} less influenced by data capture devices and environment information, {making it one of the top-performing face PADs in the recent literature. We include} this PAD solution in our experiments, considering its high PAD generalizability under cross-dataset evaluation. 

Implementation details of all solutions are discussed in Section \ref{ssec:implementation_details}.

\subsection{Evaluation metrics} % done 1st
\label{ssec:evaluation_metrics}

To measure the performance and fairness of PAD algorithms, we adopt the widely employed PAD metrics defined in the standardized ISO/IEC 30107-3 \cite{ISOPDF} and a fairness measurement metric for FR systems recently introduced in \cite{de2020fairness}, which serves as the foundation of our proposed PAD fairness metric.

\noindent \textbf{PAD performance metric:}
Following the definitions in ISO/IEC 30107-3 \cite{ISOPDF}, we use the Attack Presentation Classification Error Rate (APCER) and the {Bona fide Presentation Classification Error Rate} (BPCER) {as PAD performance metrics}. The APCER refers to the proportion of attack presentations incorrectly classified as bona fide presentations, while BPCER refers to the proportion of bona fide samples misclassified as attack samples. To cover the different operation points, we report the 1-BPCER value at six different fixed APCER values (0.5\%, 1\%, 5\%, 10\%, 15\%, 20\%) in plots. To provide a clear and straightforward comparison, we present these values in {figures} (as shown in Figure \ref{fig:bpcer_at_apcer_gender}, \ref{fig:bpcer_at_apcer_gender_ps}, \ref{fig: bpcer_at_apcer_occlusion}, \ref{fig:bpcer_at_apcer_occlusion_ps}). In addition, {we report the} Equal Error Rate (EER), which is the APCER value when APCER and BPCER are equal, to measure the overall PAD performance in Tables \ref{tab:eer_gender}, \ref{tab:eer_occlusion}, \ref{tab:eer_other}. 

\noindent \textbf{Biometric fairness metric:}
To explore the fairness of PAD performance across different attribute groups, we adapt the FDR proposed in \cite{de2020fairness}. FDR is defined to measure the fairness of a biometric verification system by leveraging a single decision threshold from all test groups.Their experimental assessments via pre-built fair and unfair systems \cite{de2020fairness} demonstrated that FDR can better represent the fairness of algorithms compared to using only ROC/DET curves. As a result, we utilize FDR to verify the fairness of PAD solutions. 
A PAD system is considered fair if different attribute groups share the same
APCER and BPCER for a given decision threshold $\tau$, where $\tau = APCER_{x}$ from all groups. Based on this theorization, FDR can be calculated as:
{\small
\begin{align}
    &A(\tau) = max(|APCER^{d_i}(\tau) - APCER^{d_j}(\tau)|), \quad \forall d_i,d_j \in D\\
    &B(\tau) = max(|BPCER^{d_i}(\tau) - BPCER^{d_j}(\tau)|), \quad \forall d_i,d_j \in D\\
    &FDR(\tau) = 1 - (\alpha A(\tau) + (1 - \alpha)B(\tau))
\end{align}
}%
where $D$ is a set of attribute groups $D = \{ d_1, d_2, ..., d_n \}$, and $\alpha$ is a hyper-parameter that defines the importance of misclassified attacks (i.e. APCER). The value of FDR varies from 0 (maximum discrepancy) to 1 (minimum discrepancy), {with smaller values indicating lower fairness.} We plot the FDR values under different decision thresholds $\tau = APCER_{x}$ where $x$ varies from 0.005 to 0.2. As stated in \cite{de2020fairness}, the FDR value of a fair system is not sensitive to {variation in $\tau$}, which reflected in the plot is that the FDR curve should be straight and {at} a higher position. 
%The $\alpha$ is set to 0.5 in our experiments as suggested in \cite{de2020fairness}, but it can be {adjusted based on the importance assigned to fairness in either APCER or BPCER for a certain application.}

\noindent \textbf{Accuracy Balanced Fairness:}
To explore the fairness of PAD performance on different attribute groups, we adapt the FDR. However, FDR does not consider absolute performance, thus a completely fair but low-performing PAD will be considered "better" than a very slightly unfair PAD that performs perfectly, according to FDR.
To further associate PAD performance and biometric fairness, we propose the Accuracy Balanced Fairness metric. Given a decision threshold $\tau$, where $\tau = APCER_{x}$ from all groups, ABF is formulated as follows:
{\small
\begin{align}
    &A(\tau) = \frac{max(|APCER^{d_i}(\tau) - APCER^{d_j}(\tau)|)} {1 - max_D( APCER(\tau))} , \quad \forall d_i,d_j \in D\\
    &B(\tau) = \frac{max(|BPCER^{d_i}(\tau) - BPCER^{d_j}(\tau)|)}{1 - max_D(BPCER(\tau))}, \quad \forall d_i,d_j \in D\\
    &ABF(\tau) = 1 - (\alpha A(\tau) + (1 - \alpha)B(\tau))
\end{align}
}%
The numerator of $A(\tau)$ and $B(\tau)$ is the same as {that of} the FDR metrics, while denominators $1 - max_D( APCER(\tau))$ and $1 - max_D( BPCER(\tau)$ are used to weight the discrepancy by considering the worst PAD cases in a set of attribute groups. As a result, a higher ABF value is linked to a smaller fairness discrepancy and a higher PAD performance. Similar to FDR values, we compute the ABF values under different decision thresholds $\tau = APCER_{x}$, where $x$ varies from 0.005 to 0.2 and {provide the} area under the ABF for overall quantitative comparison. 
{Additionally, $\alpha$ determines the relative importance of APCER and BPCER and can be adjusted depending on the specific application and the interests of the evaluation authority. We investigated the impact of varying $\alpha$ from 0.0 to 1.0 and observed that PAD solutions with high ABF values (fair) and high performance (i.e., ResNet50, DeepPixBis, and LMFD-PAD) maintain a high ABF value at different $\alpha$. Furthermore, the average ABF value computed based on different $\alpha$ values exhibited a consistent trend with that computed at $\alpha =0.5$.
Based on this observation and the suggestions in \cite{de2020fairness}, where FDR was proposed, we set $\alpha$ to 0.5 for all our experiments.}

\subsection{Implementation details}
\label{ssec:implementation_details}
In our experiments, we sampled 20 frames at equal intervals from each video (when the data format is video) for training and testing, following the common practice of frame sampling in PAD \cite{DBLP:conf/wacv/FangDKK22,deeppixbis,DBLP:conf/icb/FangAKD22}. For each frame, the face was detected and cropped using Multi-task cascaded convolutional neural network (MTCNN) \cite{DBLP:journals/spl/ZhangZLQ16} and resized to $224 \times 224 \times 3$ pixels \cite{DBLP:conf/wacv/FangDKK22,deeppixbis}. To train an LBP-MLP model, {we used} the Adam optimizer with an initial learning rate of $10^{-1}$ and binary cross-entropy loss to supervise the MLP training. 
For training the DeepPixBis \cite{deeppixbis} and LMFD-PAD \cite{DBLP:conf/wacv/FangDKK22}, we followed the implementation setups described in their respective works, which included horizontal flip, rotation, and color jitter augmentations. 
For training a ResNet model, we {used} the same training settings {as} DeepPixBis \cite{deeppixbis} with cross-entropy loss only. To address the unequal distribution of samples between attacks and bona fide, we simply {oversampled} the minority to make the ratio of bona fide and attack data close to 1:1, following the common practice \cite{DBLP:conf/wacv/FangDKK22,DBLP:journals/asc/Kovacs19,deeppixbis}. 
To further reduce overfitting, we also utilized early stopping technique with {a} maximum epochs of 100 and {a} patience of 20 epochs for ResNet \cite{deeppixbis}, DeepPixBis \cite{deeppixbis}, and LMFD-PAD \cite{DBLP:conf/wacv/FangDKK22}. The maximum training epoch {was} experimentally set to 1000 for LBP-MLP \cite{freitas2012lbp}. 
%The hyper-parameters used for training bias mitigation model were the same as used in the fairness exploration phase. The used patch size in PatchSwap solution is $64 \times 64$ and the swap probability is set to 0.5.

\section{Fairness enhancement: FairSWAP}
\label{sec:bias_mitigation}

%\subsection{Cross-attribute PatchSwap for bias mitigation} % done 1st

\begin{figure*}[htb]
\centering
\includegraphics[width=0.8\linewidth]{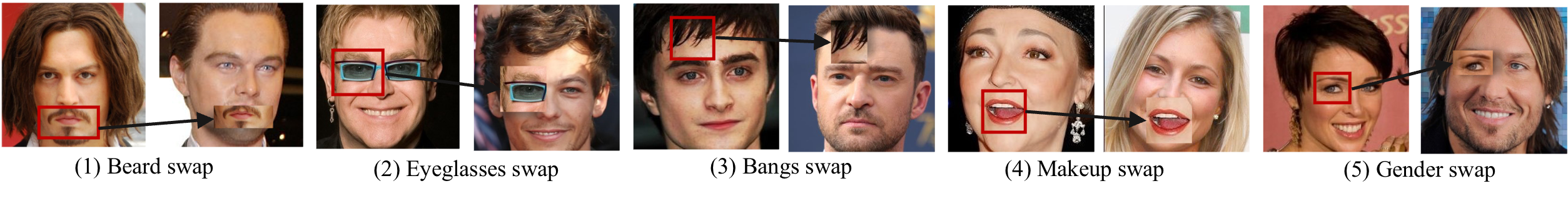}
\caption{FairSWAP: patches are swapped between training samples from different attribute groups, aiming to disorder the identity and appearance features.}
\label{fig:swapped_patches}
\end{figure*} 
In this work, four face PAD solutions are adopted to investigate the performance {and} fairness of different groups. The experimental results reported later in Section \ref{sec:fairness_assessment_baselines} indicate that all assessed PAD solutions perform unfairly on different attribute groups. For example, Table \ref{tab:eer_gender} shows that the performance on the female group is worse than on the male group when the model learns from both male and female groups. Also, Table \ref{tab:eer_other} indicates that subjects with bangs possess a worse performance than subjects without bangs. One possible reason is that face PAD algorithms may be implicitly influenced by semantic concepts related to face identity, rather than constantly focusing on discovering the {discriminative} features between bona fide and attack {samples}.
Motivated by this reasoning, we propose a fairness boosting cross-attribute patch swapping augmentation (FairSWAP) technique to mitigate the bias in trained PAD solutions.
FairSWAP swaps image patches between images of different attribute groups to disrupt the demographic/semantic information and guide PAD models to learn to differentiate between attack and bona fide clues, rather than other irrelevant clues.
{The goal of FairSWAP is to produce complex training samples with combined attributes, therefore minimizing the impact of underlying attributes unbalance on PAD models and enhancing their fairness. 
A well-performing PAD model is expected to be unaffected by non-PAD related identity and semantic information. 
Our FairSWAP helps to generate such training samples independent of the subjects, appearance, and even capture environments, based on the existing data. 
The data augmented by FairSWAP preservers discriminative attack features and obscures attribute information, and thus resulting in a fairer model.
}
%Technically similar, however attribute-oblivious, augmentation approach \cite{DBLP:conf/biosig/KantarciDE21} has showed slight enhancement in general PAD performance.
%Inspired by \cite{DBLP:conf/biosig/KantarciDE21} where the authors proposed to reconstruct face images from various small patches and we propose to swap the patches between different attribute groups to disrupt the demographic/semantic information and make sure the PAD is learning to differentiate between attack clues and bona fide clues, rather than other irrelevant clues.

As illustrated in Figure \ref{fig:swapped_patches}, given a training image, the attribute region of a subject, such as beard, eyeglasses, bangs, and makeup (e.g. lipstick), is extracted and overlaid on the sample of another training subject. In the case of gender groups, a randomly located (details to follow) region from an image of a female subject is swapped to an image of a male subject, and vice versa.
{The swapping process is implemented as a data augmentation step in which a region is randomly selected from a candidate image and swapped to the same position in another image. }
For {PAD methods that employ} pixel-wise supervision (i.e. DeepPixBis \cite{deeppixbis} and LMFD-PAD \cite{DBLP:conf/wacv/FangDKK22}), the corresponding ground truth map is updated based on the swapped locations simultaneously. 
For {PAD methods that utilize} binary supervision (i.e. LBP-MLP \cite{freitas2012lbp} and ResNet50 \cite{resnet50}), the label of the image is updated to attack if the input image after swapping contains a partial attack region; otherwise, the label of the image {remains} unchanged.
% Naser till here
The swapping strategy is detailed as follows: \\
\begin{itemize}
    \item Given a bona fide image $x_1$, a randomly located patch is swapped with a probability of $p_1$. If FairSWAP is performed, another bona fide candidate image $x_2$ (i.e. all bona fide training images except $x_1$) will be randomly selected. The training label for $x_1$ remains as bona fide.
    \item Given an attack image $x_1$, the final image fed to models is determined by three probabilities. A probability $p_2$ is first used to decide whether to apply FairSWAP. If FairSWAP is performed, $p_3$ is then used to randomly select an attack or a bona fide candidate image (i.e. all training samples excluding $x_1$). Lastly, if a bona fide image is selected, $p_4$ is used to decide the size of partial attack. The binary training labels remain as attack, while the corresponding pixel-wise map is updated according to the swapped region.
\end{itemize}

{
Considering the computational efficiency of FairSWAP, as a data augmentation technique, FairSWAP does not require extra trainable parameters and is only required during training, i.e. removed in the inference phase. The computational cost of FairSWAP is only the extra running time during data processing in the training phase. We compute the data processing time for the original steps (without FairSWAP) and the time when including FairSWAP. The average running time per image is 0.00241 seconds without FairSWAP and 0.00292 seconds with FairSWAP, computed over the entire training set.
In practice, FairSWAP requires an additional 57.42 seconds per epoch during baseline training (i.e. training on all data). 
The running time cost introduced by FairSWAP is relatively low compared to the average training time of 629.93 seconds per epoch for the three deep-learning-based models (on a single GPU RTX 2080 Ti).
Therefore, our FairSWAP is a simple yet efficient method as it does require only minimum overhead computational cost during training and no overhead computational cost during operation.}
%Moreover, the above computation of running time is based on the augmentation being performed on a single-threaded CPU, which can be optimized depending on the hardware used, and the training itself is performed on the GPU mentioned above. 
%Moreover, the above computation of running time for FairSWAP is through a single-threaded CPU, and this can be further optimized depending on the hardware used. 
%Compared to FR fairness enhancement methods that leverage attention mechanisms and additional networks \cite{DBLP:conf/cvpr/Gong0021,DBLP:conf/eccv/GongLJ20}, FairSWAP has a significantly lower running time cost. 
%FairSWAP does not add any computational cost in the inference (PAD operation) phase. Therefore, our FairSWAP is a simple yet efficient method.}

\noindent \textbf{Implementation details:} 
As FairSWAP serves as a data augmentation technique and can be plugged into any training process, {we use the same} image pre-processing and hyper-parameters (e.g., optimizer and learning rate) {as those employed in the fairness assessment} in Section \ref{ssec:implementation_details} for training the fairness enhancement model. The swapped patch size in FairSWAP is $64 \times 64$ in our experiments and the probabilities $p_1$, $p_2$, $p_3$, and $p_4$ are manually set to 0.3, 0.3, 0.5, and 0.5, respectively, based on empirically experiments. 

\section{Results of fairness assessment} % ready for Naser
\label{sec:fairness_assessment_baselines}
This section presents the PAD performance across attribute groups in terms of detection EER and assesses the fairness under various ODTAs in terms of FDR and {the} proposed ABF.
To explore the data-induced potential bias, we adopt four PAD models described in Section \ref{ssec:pad_algorithms} on different data groups following the experimental protocols presented in Section \ref{ssec:experimental_protocols}. The overall PAD performance is reported in terms of EER values in Table \ref{tab:eer_gender}, \ref{tab:eer_occlusion}, and \ref{tab:eer_other}. In addition, 1-BPCER values are computed based on different ODTAs and illustrated in Figure \ref{fig:bpcer_at_apcer_gender}, \ref{fig: bpcer_at_apcer_occlusion}. In the following, we discuss the results over gender groups, occlusion and non-occlusion groups, and the remaining attribute groups.

\subsection{Fairness assessment over gender (Protocol-1)}
\label{ssec:results_gender_baselines}

%====================================================================
\begin{table*}[thb]
\centering
\def\arraystretch{0.8}
\resizebox{0.9\textwidth}{!}{
\begin{tabular}{cc|ccc|ccc|ccc|ccc}
\hline
\multirow{2}{*}{Trained}   &  \multirow{2}{*}{Test}        & \multicolumn{3}{c|}{LBP \cite{freitas2012lbp}}      & \multicolumn{3}{c|}{ResNet50 \cite{resnet50}} & \multicolumn{3}{c|}{DeepPixBis \cite{deeppixbis}} & \multicolumn{3}{c}{LMFD-PAD \cite{DBLP:conf/wacv/FangDKK22}} \\ %\hline \hline
&  & B   & FairSwap        & Impro & B     & FairSwap    & Impro   & B            & FairSwap & Impro & B     & FairSwap   & Impro   \\ \hline
\multirow{2}{*}{Fused} &
  M &
  11.13 &
  11.27 &
  +0.14 &
  2.54 &
  1.86 &
  -0.68 &
  1.17 &
  1.03 &
  -0.14 &
  1.94 &
  1.72 &
  -0.22 \\
 &
  F &
  \textbf{17.69} &
  \textbf{14.06} &
  -3.63 &
  \textbf{3.00} &
  \textbf{2.34} &
  -0.66 &
  \textbf{1.57} &
  \textbf{1.29} &
  -0.28 &
  \textbf{2.62} &
  \textbf{2.25} &
  -0.37 \\ \hline
\multirow{2}{*}{M} &
  M &
  12.55 &
  12.88 &
  +0.33 &
  2.96 &
  1.54 &
  -1.42 &
  1.32 &
  1.45 &
  +0.13 &
  2.15 &
  2.68 &
  +0.53 \\
 &
  F &
  \textbf{16.17} &
  \textbf{17.12} &
  +0.95 &
  \textbf{5.90} &
  \textbf{2.34} &
  -3.56 &
  \textbf{3.50} &
  \textbf{2.56} &
  -0.94 &
  \textbf{3.89} &
  \textbf{3.87} &
  -0.02 \\ \hline
\multirow{2}{*}{F} &
  M &
  \textbf{19.04} &
  \textbf{22.48} &
  +3.44 &
  \textbf{13.13} &
  \textbf{9.13} &
  -4.00 &
  \textbf{10.95} &
  \textbf{11.77} &
  +0.82 &
  \textbf{9.88} &
  \textbf{10.85} &
  +0.97 \\
 &
  F &
  18.67 &
  21.98 &
  +3.31 &
  10.62 &
  8.25 &
  -2.37 &
  7.52 &
  8.42 &
  +0.90 &
  9.15 &
  8.33 &
  -0.82 \\ \hline
\end{tabular}}
\caption{PAD performance in terms of EER  (\%) on gender groups by baseline models (B) and FairSWAP. Fused, M, and F refer to the fused male and female data, male, and female data, respectively. Impro refers to percentage improvement. The results are obtained by separately training models on fused, male, and female training data of CAAD-PAD. Bold numbers indicate the highest EER values between male and female test data by each trained model.}
\label{tab:eer_gender}
\vspace{-5mm}
\end{table*}
%=====================================================================

%=====================================================================
\begin{figure*}[th]
\centering
\foreach \model in {baseline,male,female}{
    \foreach \db in {LBP,DeepPixBis,ResNet50,LMFD}{
        \includegraphics[width=0.23\linewidth]{gender_No_PS_\db_\model.pdf}
    }
}
\includegraphics[width=0.8\linewidth]{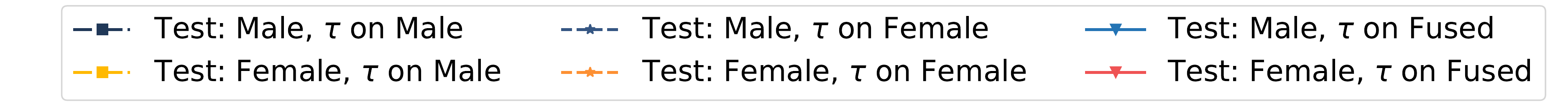}
\caption{The results of baseline models on gender groups in terms of $BPCER\ @\ APCER_x$. To study the impact of ODTA on fairness, the threshold $\tau$ is calculated separately on fused, male, and female test set. When evaluating models trained on fused data, using a threshold computed on male samples shows a relatively higher performance than a threshold from fused or male groups.}
\label{fig:bpcer_at_apcer_gender}
\vspace{-3mm}
\end{figure*}
%=====================================================================

\noindent \textbf{PAD Performance with respect to gender:}
The experiments were conducted following our Protocol-1 (Section \ref{ssec:experimental_protocols}) to assess the differential performance and outcome, and fairness on gender groups. Four PAD models were trained separately on the entire training set (Fused), males (M), and females (F) of CAAD-PAD. The results were then reported separately on male and female test data in Table \ref{tab:eer_gender}. 
The EER values in columns 'B' refer to baseline results (results of FairSWAP will be discussed later in Section \ref{sec:bias_mitigation_PS}) and {the} bold numbers indicate higher error rates between male and female groups. 
In the case of models trained separately on male and female samples, the error rates are higher for the gender group {that is unseen during training.} Such results are reasonable as it is a challenge for models to generalize on data with unknown aspects. 
When models are trained on the fused data, the male test set obtains consistently lower EER values than females, indicating that the male group possesses a relatively higher protection from PAD systems. Note that the ratio of male and female samples in the fused training data is close to 1:1, achieved by data oversampling (details Section \ref{ssec:implementation_details}).
Moreover, models trained on the male group achieve consistently lower EER values on both test sets than models trained on the female group. For example, when models are trained on males, DeepPixBis \cite{deeppixbis} and LMFD-PAD \cite{DBLP:conf/wacv/FangDKK22} achieve the lowest EER value of 3.50\% and 3.89\%, respectively, while the lowest EER values obtained by models trained on females are 10.95\% by DeepPixBis and 9.85\% by LMFD-PAD. Both observations suggest that a model tends to learn a better feature representation from the male group and shows a better PAD generalizability on the male group, which is consistent with the observations in FR \cite{DBLP:journals/tifs/KlareBKBJ12,DBLP:conf/cvpr/WangD20}.
% might add reason, like makep, bangs 

Figure \ref{fig:bpcer_at_apcer_gender} explores the PAD performance under different ODTAs by illustrating the 1-BPCER($\tau$) values at thresholds $\tau = APCER_x$ separately computed from fused, male, and female groups. 
By observing Figure \ref{fig:bpcer_at_apcer_gender}, we conclude that 1) When evaluating models trained on fused data, using a decision threshold computed on female samples results in a relatively lower performance than a threshold from fused or male groups. 2) When evaluating models trained on males, the PAD performance is consistently higher (i.e. higher 1-BPCER values) on male test data than on females, irrespective of ODTAs. 3) However, when evaluating models trained on females, male test data obtains comparable (by ResNet50 and DeepPixBis) and even higher PAD performance (by LBP-MLP) than testing on females, unlike the previous EER results observation that demographic data known in the training set exhibits better PAD performance. 
These observations indicate that in addition to the bias caused by the nature of training data, the ODTA is also one of the triggers of PAD unfairness.

%=====================================================================
% gender fairness assessment 
\begin{figure*}[thb]
\centering
\includegraphics[width=0.9\linewidth]{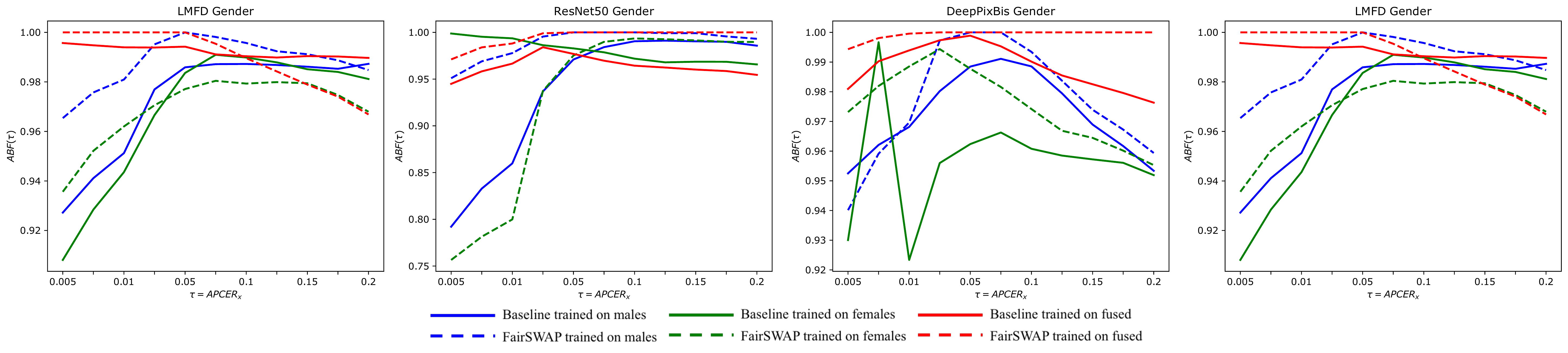}
\vspace{-3mm}
\caption{Fairness measure in terms of ABF values by employing four face PAD models on gender groups. Solid lines represents the results of baseline models, dashed lines represents the results of FairSWAP. Higher and smoother lines indicate higher fairness. Models trained on separate gender groups possess more discrepancies than the same PAD models trained on fused data.}
\label{fig:abf_gender}
\end{figure*} 

\begin{table*}[th]
\centering
\def\arraystretch{0.8}
\resizebox{0.9\textwidth}{!}{
\begin{tabular}{c|l|cc|cc|cc|cc}
\hline
\multirow{2}{*}{{Metric}} & \multirow{2}{*}{Trained}     & \multicolumn{2}{c|}{LBP \cite{freitas2012lbp}}      & \multicolumn{2}{c|}{ResNet50 \cite{resnet50}} & \multicolumn{2}{c|}{DeepPixBis \cite{deeppixbis}} & \multicolumn{2}{c}{LMFD-PAD \cite{DBLP:conf/wacv/FangDKK22}} \\
&  & B  & FairSwap & B   & FairSwap  & B   & FairSwap & B  & FairSwap  \\ \hline
%& FDR & FDR & FDR & FDR & FDR & FDR & FDR & FDR  \\ \hline[dashed]
\multirow{4}{*}{\rotatebox[origin=c]{90}{FDR}} & Fused & 0.821 & \textbf{0.870} & 0.882 & \textbf{0.900} & 0.900 & \textbf{0.904} & \textbf{0.902} & 0.895 \\
& Male & 0.775 & \textbf{0.820} & 0.871 & \textbf{0.895} & 0.886 & \textbf{0.888} & 0.883 & \textbf{0.889} \\
& Female & 0.871 & \textbf{0.901} & \textbf{0.893} & 0.874 & 0.881 & \textbf{0.885} & \textbf{0.889} & 0.882 \\ \cline{2-10} %\hline[dashed]
& Average & 0.822 & \textbf{0.864} &	0.882 & \textbf{0.890} & 0.889 & \textbf{0.892} & \textbf{0.891} & 0.889 \\ \hline \hline
\multirow{4}{*}{\rotatebox[origin=c]{90}{ABF}} & Fused  & 0.695  & \textbf{0.762}  & 0.877 & \textbf{0.905} & 0.899  & \textbf{0.909} & \textbf{0.901}  & \textbf{0.901}  \\
& Male          & \textbf{0.443}       & 0.358                & 0.857                & \textbf{0.901}       & 0.885 & \textbf{0.890} & 0.885  & \textbf{0.899} \\
& Female  & 0.723& \textbf{0.894} & \textbf{0.891} & 0.848  & 0.871  & \textbf{0.888} & 0.882  & \textbf{0.883} \\ \cline{2-10}
& Average & 0.620 & \textbf{0.671} & 	0.875 &	\textbf{0.885} & 0.885&	\textbf{0.896} &	0.889 &	\textbf{0.894}\\ \hline
\end{tabular}}
\caption{Fairness in terms of FDR-AUC and ABF-AUC with respect to gender. The higher FDR-AUC and ABF-AUC indicate a fairer PAD (highest in bold). FairSWAP enhances the fairness in most cases.}
\label{tab:fdr_abf_auc_gender}
\vspace{-3mm}
\end{table*}
%=====================================================================

\noindent \textbf{Fairness with respect to gender:}
%The above analysis explores differential performance and outcome of PAD induced by training data and ODTAs. To further assess the training data-related fairness and performance in PAD, we illustrate ABFs in Figure \ref{fig:abf_gender}.
{Figure \ref{fig:abf_gender} shows the training data-related fairness nad performance in PAD by plotting ABFs.} 
Additionally, Table \ref{tab:fdr_abf_auc_gender} presents the area under FDR (FDR-AUC) and area under ABF (ABF-AUC) of each PAD system. 
As suggested in \cite{de2020fairness}, a more stable and higher FDR curve and a higher FDR-AUC indicate a fairer system. 
{As shown in Figure \ref{fig:abf_gender}, all solid ABF curves, representing the fairness of baseline models, exhibited varying degrees of fluctuation. }
Moreover, deep-learning-based models trained on separate gender groups (blue and green curves) have more discrepancies than the same PAD model trained on fused data (red curves) in most cases{, indicating that the nature of training data distribution does affect the fairness of PADs.}
Such observations are consistent with the ABF-AUC values in Table \ref{tab:fdr_abf_auc_gender}. 
{As shown in Table \ref{tab:fdr_abf_auc_gender}, the use of ABF metric significantly enlarges the numerical difference between deep learning and hand-crafted feature-based models, in comparison to FDR metric. This finding confirms the effectiveness of our ABF metric in jointly considering both absolute PAD performance and fairness, thus preventing the uninformed preference of PAD model that performs poorly but fairly.}

%=====================================================================
\begin{table*}[th]
\centering
\def\arraystretch{0.8}
\resizebox{0.9\textwidth}{!}{
\begin{tabular}{cc|ccc|ccc|ccc|ccc}
\hline
\multirow{2}{*}{Trained}   &  \multirow{2}{*}{Test}        & \multicolumn{3}{c|}{LBP \cite{freitas2012lbp}}      & \multicolumn{3}{c|}{ResNet50 \cite{resnet50}} & \multicolumn{3}{c|}{DeepPixBis \cite{deeppixbis}} & \multicolumn{3}{c}{LMFD-PAD \cite{DBLP:conf/wacv/FangDKK22}} \\ %\hline \hline
&  & B   & FairSwap        & Impro & B     & FairSwap    & Impro   & B            & FairSwap & Impro & B     & FairSwap   & Impro   \\ \hline
\multirow{2}{*}{Fused} &
  O &
  \textbf{13.40} &
  11.61 &
  -1.79 &
  \textbf{2.93} &
  \textbf{2.65} &
  -0.28 &
  1.19 &
  \textbf{1.46} &
  +0.27 &
  \textbf{2.54} &
  \textbf{2.28} &
  -0.26 \\
                   & w/o O & 13.29 & \textbf{12.4} & -0.89 & 2.04    & 1.82      & -0.22   & \textbf{1.33}  & 0.95   & -0.38 & 1.98    & 1.67     & -0.31   \\ \hline
\multirow{2}{*}{O} & O     & 11.11 & 15.05         & +3.94 & 3.64    & 2.68      & -0.96   & 1.21           & 1.87   & +0.66 & 2.87    & 2.80      & -0.07   \\
 &
  w/o O &
  \textbf{14.71} &
  \textbf{18.3} &
  +3.59 &
  \textbf{6.27} &
  \textbf{2.83} &
  -3.44 &
  \textbf{2.69} &
  \textbf{3.43} &
  +0.74 &
  \textbf{3.21} &
  \textbf{3.22} &
  +0.01 \\ \hline
\multirow{2}{*}{w/o O} &
  O &
  \textbf{22.60} &
  \textbf{21.43} &
  -1.17 &
  \textbf{11.40} &
  \textbf{7.89} &
  -3.51 &
  \textbf{9.91} &
  \textbf{6.62} &
  -3.29 &
  \textbf{8.94} &
  \textbf{8.19} &
  -0.75 \\
                   & w/o O & 19.46 & 19.60          & +0.14 & 4.94    & 4.31      & -0.63   & 6.08           & 3.86   & -2.22 & 4.82    & 4.68     & -0.14   \\ \hline
\end{tabular}}
\caption{PAD performance in terms of EER (\%) on occlusion groups by baseline models (B) and FairSWAP. Fused, O, and w/o O refer to the fused data, occlusion, and non-occlusion data, respectively. Bold numbers indicate the highest EER values between occlusion and non-occlusion group by each model. The results imply that occlusion group is more challenging to be correctly classified than non-occlusion group, but can help models learn more complex and generalized representations.}
\label{tab:eer_occlusion}
\vspace{-3mm}
\end{table*}
%=====================================================================

\subsection{Fairness assessment over occlusion (Protocol-2)}
\label{ssec:results_occlusion_baselines}

%=====================================================================
\begin{figure*}[th]
\centering
\foreach \model in {baseline,occlusion,noocclusion}{
    \foreach \db in {LBP,DeepPixBis,ResNet50,LMFD}{
        \includegraphics[width=0.23\linewidth]{occ_No_PS_\db_\model.pdf}
    }
}
\includegraphics[width=0.8\linewidth]{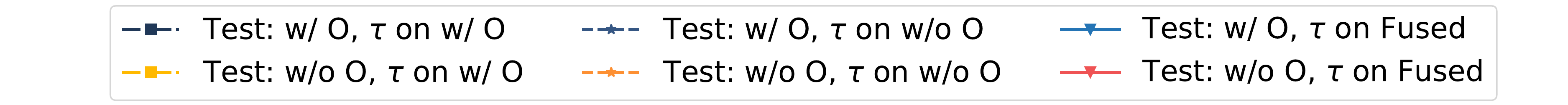}
\caption{The results of baseline models on occlusion group in terms of $BPCER\ @\ APCER_x$. To study the impact of ODTA on fairness, the threshold $\tau$ is calculated separately on fused, occlusion, and non-occlusion test set. Performance of models trained on fused data degrades when assigning a $\tau$ driven from the occlusion group. }
\label{fig: bpcer_at_apcer_occlusion}
\end{figure*}
%=====================================================================

\begin{figure*}[htb]
\centering
\includegraphics[width=0.9\linewidth]{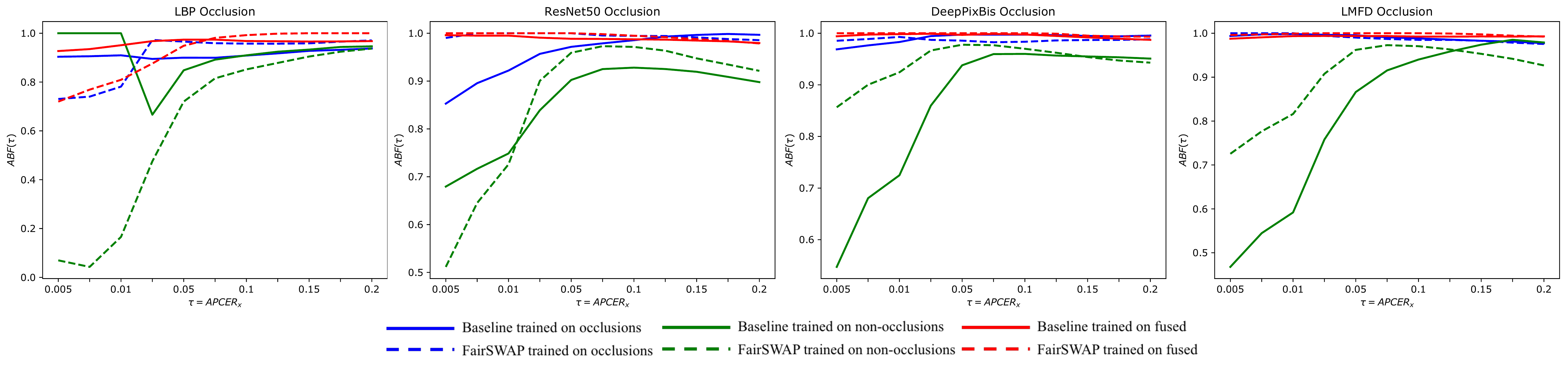}
\caption{Fairness measure in terms of ABF values by employing four face PAD models on occlusion groups. Solid lines represents the results of baseline models, dashed lines represents the results of FairSWAP. Higher and smoother lines indicate higher fairness. Models trained on fused and occluded data exhibit higher fairness.}
\label{fig:abf_occlusion}
\vspace{-3mm}
\end{figure*}

%To assess the occlusion-related fairness of face PAD model, we followed Protocol-2 to demonstrate the experiments on three different training set, fused set (denotes as Fused), occlusion set (O), and without occlusion set (w/ O). In our case, subjects wearing eyeglasses, or wearing beard, or having hair bangs are categorized as occlusion data. 
We report the face PAD performance in terms of EER values on occlusion/non-occlusion group in Table \ref{tab:eer_occlusion} and the 1-BPCER($\tau$) value at different ODTAs in Figure \ref{fig: bpcer_at_apcer_occlusion}. Furthermore, we assess the fairness of PAD models trained on various composition of training data in Figure \ref{fig:abf_occlusion} and Table \ref{tab:fdr_abf_auc_occlusion}.

\noindent \textbf{PAD Performance with respect to occlusion:}
{From} Table \ref{tab:eer_occlusion}, we observe that: 1) {Models trained on separate occlusion and non-occlusion data perform better on their respective learned test groups.} 2) Models trained on non-occluded samples obtain significantly higher EER values than those trained on occluded samples for both test groups. 3) Models trained on fused data achieve slightly higher EER values on occlusion group in most cases. 
This suggests that occluded data is harder to be classified correctly than non-occluded data, but it can help models to learn more complex and generalized representations. 
Furthermore, Figure \ref{fig: bpcer_at_apcer_occlusion} shows that the performance of models trained on fused data is decreased when using decision thresholds $\tau$ {drived} from occluded data. 
In addition, when $\tau$ at smaller APCER, LBP-MLP and DeepPixBis models perform worse on the non-occluded group than on the occluded group (i.e. curves of orange series lower than curves of blue series), which is in contrast to previous observations in terms of EER values. 
Despite some performance degradation of models trained on separate group data by various thresholds, the tendencies of differential outcomes between two test groups remain coincident with the observations in terms of EER values. Therefore, we conclude that the nature of training data and ODTAs are triggers of the performance bias in PAD. 

\begin{table*}[thbp!]
\centering
\def\arraystretch{0.8}
\resizebox{0.9\textwidth}{!}{
\begin{tabular}{c|l|cc|cc|cc|cc}
\hline
\multirow{2}{*}{{Metric}} & \multirow{2}{*}{Trained}     & \multicolumn{2}{c|}{LBP \cite{freitas2012lbp}}      & \multicolumn{2}{c|}{ResNet50 \cite{resnet50}} & \multicolumn{2}{c|}{DeepPixBis \cite{deeppixbis}} & \multicolumn{2}{c}{LMFD-PAD \cite{DBLP:conf/wacv/FangDKK22}} \\
&  & B  & FairSwap & B   & FairSwap  & B   & FairSwap & B  & FairSwap  \\ \hline
\multirow{4}{*}{\rotatebox[origin=c]{90}{FDR}} & Fused & 0.887 & \textbf{0.892} & 0.898 & \textbf{0.899} & 0.904 & \textbf{0.905} & 0.901 & \textbf{0.902} \\
& Occlusion & 0.856 & \textbf{0.884} & 0.883 & \textbf{0.899} & \textbf{0.901} & 0.891 & \textbf{0.901} & 0.894 \\
& w/o Occlusion & \textbf{0.875} & 0.841 & 0.832 & \textbf{0.834} & 0.844 & \textbf{0.865} & 0.815 & \textbf{0.849} \\   \cline{2-10}
& Average & \textbf{0.873} &	0.872 &	0.871 &	\textbf{0.877} &	0.883 &	\textbf{0.887} & 0.872 & \textbf{0.882} \\ \hline \hline
\multirow{4}{*}{\rotatebox[origin=c]{90}{ABF}} & Fused      & \textbf{0.874}       & 0.839                & 0.898                & \textbf{0.904}       & 0.904                & \textbf{0.908}       & 0.902                & \textbf{0.907}       \\
& Occlusion     & 0.828                & \textbf{0.830}       & 0.874                & \textbf{0.905}       & \textbf{0.901}       & 0.897                & 0.900                & \textbf{0.901}       \\
& w/o occlusion & \textbf{0.826}       & 0.571                & 0.781                & \textbf{0.794}       & 0.794                & \textbf{0.861}       & 0.751                & \textbf{0.826} \\  \cline{2-10}
& Average & \textbf{0.843} & 0.747 & 0.851	& \textbf{0.868}	& 0.866	& \textbf{0.889}	& 0.851	& \textbf{0.878} \\ \hline
\end{tabular}}
\caption{Fairness in terms of FDR-AUC and ABF-AUC with respect to occlusion attribute. The higher FDR-AUC and ABF-AUC indicate a fairer PAD model (highest in bold). FairSWAP boosts the average fairness of PAD models in most cases.}
\label{tab:fdr_abf_auc_occlusion}
\vspace{-3mm}
\end{table*}

%=====================================================================
%=====================================================================
\noindent \textbf{Fairness with respect to occlusion:}
Figure \ref{fig:abf_occlusion} illustrates the ABF values, indicating that deep-learning-based models trained on the non-occlusion group are significantly unfairer (solid green curves) than models trained on the occlusion group and fused data group (solid blue and red curves). 
{When models trained on fused data, the models exhibit} smaller fairness discrepancies, i.e. higher and {more stable} ABF curves. 
{The} ABF-AUC values in Table \ref{tab:fdr_abf_auc_occlusion} support these findings, {with models trained on fused data achieve the highest ABF-AUC values. These results confirm that diverse training data can results in better generalized and fairer PAD models.} 

\subsection{Fairness assessment over other attributes (Protocol-3)}
\label{ssec:results_attribute_baselines}
This section investigates the PAD performance and fairness on six pairwise non-demograpihc attributes: with/without bangs, beard, eyeglasses, makeup, long hair, and straight hair. We omit the assessment of the impact of training data due to insufficient data across such attributes. 
The PAD performances on these attributes are evaluated by adopting the models trained on the entire training set of CAAD-PAD. 

%=====================================================================
\begin{table*}[ht!]
\centering
\def\arraystretch{0.8}
\resizebox{0.9\textwidth}{!}{
\begin{tabular}{lc|ccc|ccc|ccc|ccc}
\hline
\multirow{2}{*}{Attribute}   &  \multirow{2}{*}{Type}        & \multicolumn{3}{c|}{LBP \cite{freitas2012lbp}}      & \multicolumn{3}{c|}{ResNet50 \cite{resnet50}} & \multicolumn{3}{c|}{DeepPixBis \cite{deeppixbis}} & \multicolumn{3}{c}{LMFD-PAD \cite{DBLP:conf/wacv/FangDKK22}} \\ %\hline \hline
&  & B   & FairSwap        & Impro & B     & FairSwap    & Impro   & B            & FairSwap & Impro & B    & FairSwap   & Impro   \\ \hline
\multirow{2}{*}{Bangs} &
  w &
  \textbf{13.42} &
  \textbf{12.70} &
  -0.72 &
  \textbf{3.71} &
  \textbf{3.49} &
  -0.22 &
  \textbf{1.36} &
  \textbf{1.60} &
  +0.24 &
  \textbf{2.64} &
  \textbf{2.54} &
  -0.10 \\
 &
  w/o &
  13.19 &
  12.05 &
  -1.14 &
  2.31 &
  1.54 &
  -0.77 &
  1.27 &
  0.92 &
  -0.35 &
  1.91 &
  1.57 &
  -0.34 \\ \hline
\multirow{2}{*}{Beard} &
  w &
  12.16 &
  8.21 &
  -3.95 &
  \textbf{2.96} &
  \textbf{2.08} &
  -0.88 &
  0.42 &
  0.46 &
  +0.04 &
  1.98 &
  1.78 &
  -0.20 \\
 &
  w/o &
  \textbf{13.33} &
  \textbf{12.39} &
  -0.94 &
  2.65 &
  2.05 &
  -0.60 &
  \textbf{1.38} &
  \textbf{1.16} &
  -0.22 &
  \textbf{2.17} &
  \textbf{1.85} &
  -0.32 \\ \hline
\multirow{2}{*}{Eye glass} &
  w &
  \textbf{13.61} &
  11.41 &
  -2.20 &
  \textbf{3.25} &
  \textbf{2.28} &
  -0.97 &
  1.25 &
  \textbf{1.61} &
  +0.36 &
  \textbf{2.61} &
  \textbf{2.25} &
  -0.36 \\
 &
  w/o &
  13.45 &
  \textbf{12.87} &
  -0.58 &
  1.96 &
  1.86 &
  -0.10 &
  \textbf{1.31} &
  0.67 &
  -0.64 &
  1.76 &
  1.44 &
  -0.32 \\ \hline
\multirow{2}{*}{Makeup} &
  w &
  12.20 &
  \textbf{14.12} &
  +1.92 &
  \textbf{2.69} &
  1.34 &
  -1.35 &
  \textbf{2.49} &
  0.93 &
  -1.56 &
  \textbf{4.16} &
  1.19 &
  -2.97 \\
 &
  w/o &
  \textbf{13.33} &
  11.95 &
  -1.38 &
  2.68 &
  \textbf{2.12} &
  -0.56 &
  1.22 &
  \textbf{1.11} &
  -0.11 &
  1.99 &
  \textbf{1.76} &
  -0.23 \\ \hline
\multirow{2}{*}{LongHair} &
  w &
  \textbf{21.69} &
  \textbf{16.80} &
  -4.89 &
  2.50 &
  \textbf{2.88} &
  +0.38 &
  \textbf{1.82} &
  \textbf{1.84} &
  +0.02 &
  \textbf{3.44} &
  \textbf{2.84} &
  -0.60 \\
 &
  w/o &
  11.11 &
  11.10 &
  -0.01 &
  \textbf{2.67} &
  1.67 &
  -1.00 &
  1.19 &
  1.05 &
  -0.14 &
  1.80 &
  1.61 &
  -0.19 \\ \hline
\multirow{2}{*}{StraightHair} &
  w &
  12.90 &
  12.09 &
  -0.81 &
  \textbf{2.82} &
  \textbf{2.17} &
  -0.65 &
  \textbf{1.38} &
  \textbf{1.24} &
  -0.14 &
  \textbf{2.30} &
  \textbf{1.92} &
  -0.38 \\
 &
  w/o &
  \textbf{14.36} &
  \textbf{12.81} &
  -1.55 &
  1.68 &
  1.19 &
  -0.49 &
  0.67 &
  0.09 &
  -0.58 &
  0.83 &
  0.97 &
  +0.14 \\ \hline \hline 
Average & & 13.73 & 12.38& 	-1.35 &  2.66 & 	2.06& 	-0.60& 	1.31& 	1.06& 	-0.26& 	2.30& 	1.81& 	-0.49 \\ \hline
\end{tabular}}
\caption{PAD performance  in terms of EER (\%) on the other attribute groups by baseline (B) and FairSWAP models trained on fused data. Type w and w/o refer to test samples with such attribute, and without such attribute, respectively. Bold numbers indicate the highest EER values between w and w/o test attribute group by each trained model. Non-demographic attributes exhibit an implicit effect on PAD to some extent.}
\label{tab:eer_other}
\vspace{-3mm}
\end{table*}
%=====================================================================

\noindent \textbf{PAD Performance with respect to other attributes:}
Table \ref{tab:eer_other} presents the {baseline PAD performance in column "B", with bold numbers indicating the highest EER value between each paired attribute for each PAD approach.} 
{The detection of samples carrying eyeglasses, bangs, and makeup is more challenging compared to those without these attributes. Conversely, subjects with beards tend to exhibit higher protection from PAD models than those without beards in most cases. This observation implicitly reveals that PAD models perform better on the male group than the female group, consistent with findings in Section \ref{ssec:results_gender_baselines}.}
{Samples with long and straight hair} achieve higher error rates than samples without these attributes in most cases. This may be attributed to the higher possibility of long hair covering the ear region and thus leading to the loss of discriminative patterns, based on empirical observations of cropped faces.
Additionally, PAD models might learn the pattern of curly hair somehow more straightforward than straight hair. Overall, {these attributes are interrelated and entangled, making it challenging to analyze the reasons behind the PAD behavior in a stand-alone manner.} In conclusion, non-demographic attributes somehow also exhibit implicit influences on PAD performance.

%EER values of samples with bangs are consistently higher than {those} without bangs for all training models, while subjects wearing eyeglasses are more challenging to detect correctly. For {the} attribute \textit{Beard}, a different behaviour is observed in comparison with the other two occlusion annotations (\textit{Bangs} and \textit{Eyeglasses}), i.e. models perform better on samples with beards in most cases. Similarly, samples without makeup obtain lower EER values in most cases. This may be attributed by that \textit{Beard} attribute is correlate with the male group.
%=====================================================================
\begin{table*}[thbp!]
\centering
\def\arraystretch{0.8}
\resizebox{0.9\textwidth}{!}{
\begin{tabular}{c|l|cc|cc|cc|cc}
\hline
\multirow{2}{*}{{Metric}} & \multirow{2}{*}{Tested}     & \multicolumn{2}{c|}{LBP \cite{freitas2012lbp}}      & \multicolumn{2}{c|}{ResNet50 \cite{resnet50}} & \multicolumn{2}{c|}{DeepPixBis \cite{deeppixbis}} & \multicolumn{2}{c}{LMFD-PAD \cite{DBLP:conf/wacv/FangDKK22}} \\
&  & B  & FairSwap & B   & FairSwap  & B   & FairSwap & B  & FairSwap  \\ \hline
\multirow{6}{*}{\rotatebox[origin=c]{90}{FDR}} & Bangs        & {0.893} & {0.895} & {0.900} & 0.890          & 0.901          & {0.901} & 0.895          & {0.899} \\
& Beard        & 0.832          & 0.834          & {0.889} & {0.902} & 0.890          & 0.898          & {0.901} & 0.897          \\
& Eyeglasses   & {0.883} & {0.899} & 0.884          & 0.895          & 0.899          & 0.898          & {0.902} & {0.902} \\
& Makeup       & 0.859          & 0.859          & 0.870          & {0.901} & 0.893          & {0.901} & 0.893          & 0.887          \\
& LongHair     & 0.814          & 0.864          & 0.878          & 0.894          & {0.902} & 0.897          & 0.895          & 0.889          \\
& StraightHair & 0.850          & 0.882          & 0.877          & 0.889          & {0.905} & {0.901} & 0.885          & 0.896          \\  \cline{2-10}
& Average      & 0.855          & \textbf{0.872} & 0.883          & \textbf{0.895} & 0.898          & \textbf{0.899} &  \textbf{0.895}          & \textbf{0.895} \\ \hline \hline
\multirow{6}{*}{\rotatebox[origin=c]{90}{ABF}} & Bangs        & {0.884} & {0.837} & {0.900} & 0.895          & 0.900          & {0.906} & 0.893          & {0.905} \\
& Beard        & 0.779          & 0.647          & {0.886} & {0.907} & 0.887          & 0.904          & {0.900} & 0.902          \\
& Eyeglasses   & {0.872} & {0.879} & 0.881          & 0.899          & 0.898          & 0.902          & {0.902} & {0.907} \\
& Makeup       & 0.743          & 0.399$*$          & 0.858          & {0.906} & 0.891          & {0.906} & 0.891          & 0.892          \\
& LongHair     & 0.700          & 0.795          & 0.872          & 0.901          & {0.901} & 0.903          & 0.893          & 0.894          \\
& StraightHair & 0.803          & 0.816          & 0.873          & 0.894          & {0.905} & 0.894          & 0.893          & 0.902          \\  \cline{2-10} 
& Average      & \textbf{0.797} & 0.729          & 0.878          & \textbf{0.900} & 0.897          & \textbf{0.903} & 0.895          & \textbf{0.900}  \\ \hline 
\end{tabular}}
\caption{Fairness in terms of FDR-AUC and ABF-AUC with respect to other attributes by using models trained on fused data. The higher FDR-AUC and ABF-AUC in bold indicate a fairer PAD model. FairSWAP enhances the fairness of all deep-learning-based PAD models.}
\label{tab:fdr_abf_auc_attribute}
\vspace{-5mm}
\end{table*}
%=====================================================================

\noindent \textbf{Fairness with respect to other attributes:}
Table \ref{tab:fdr_abf_auc_attribute} presents {the} FDR-AUC and ABF-AUC values. We observe {the following}: 1) Occlusion attribute groups (bangs, beard and eyeglasses) achieve the highest FDR-AUC and ABF-AUC values in most cases (six out of eight baseline cases), {indicating their higher fairness compared to non-occlusion group.} 2) Deep-learning-based PAD solutions are fairer than the hand-crafted feature-based approach, especially when considering ABF values. These results may be attributed to the fact that deep learning models learn more abstract features beyond texture features. 
Overall, the fairness of PAD models slightly varies across non-demographic attribute groups, {This variation may be caused by inherent differences in the numbers of training samples in each attribute group.}
%Overall, even with the diverse training data, PAD models exhibit the slightly differential fairness on the varied test attribute groups, which is in accord with studies in \cite{de2020fairness} that test data is also a potential induction of bias.

\section{Results of fairness enhancement}
\label{sec:bias_mitigation_PS}
The fairness investigation results on CAAD-PAD ( Section \ref{sec:fairness_assessment_baselines}) revealed that the training data distribution and ODTA are triggers of PAD unfairness. This may be attributed to {the} PAD models learning  identity and semantic information beyond the discriminative attack clues. To enhance PAD fairness, we proposed the FairSWAP method, acting as a data augmentation technique in Section \ref{sec:bias_mitigation}. This section discusses {the} fairness enhancement results obtained by applying the FairSWAP solution.

\subsection{Fairness enhancement over gender (Protocol-1)}
\label{ssec:ps_gender_results}
\begin{figure*}[th]
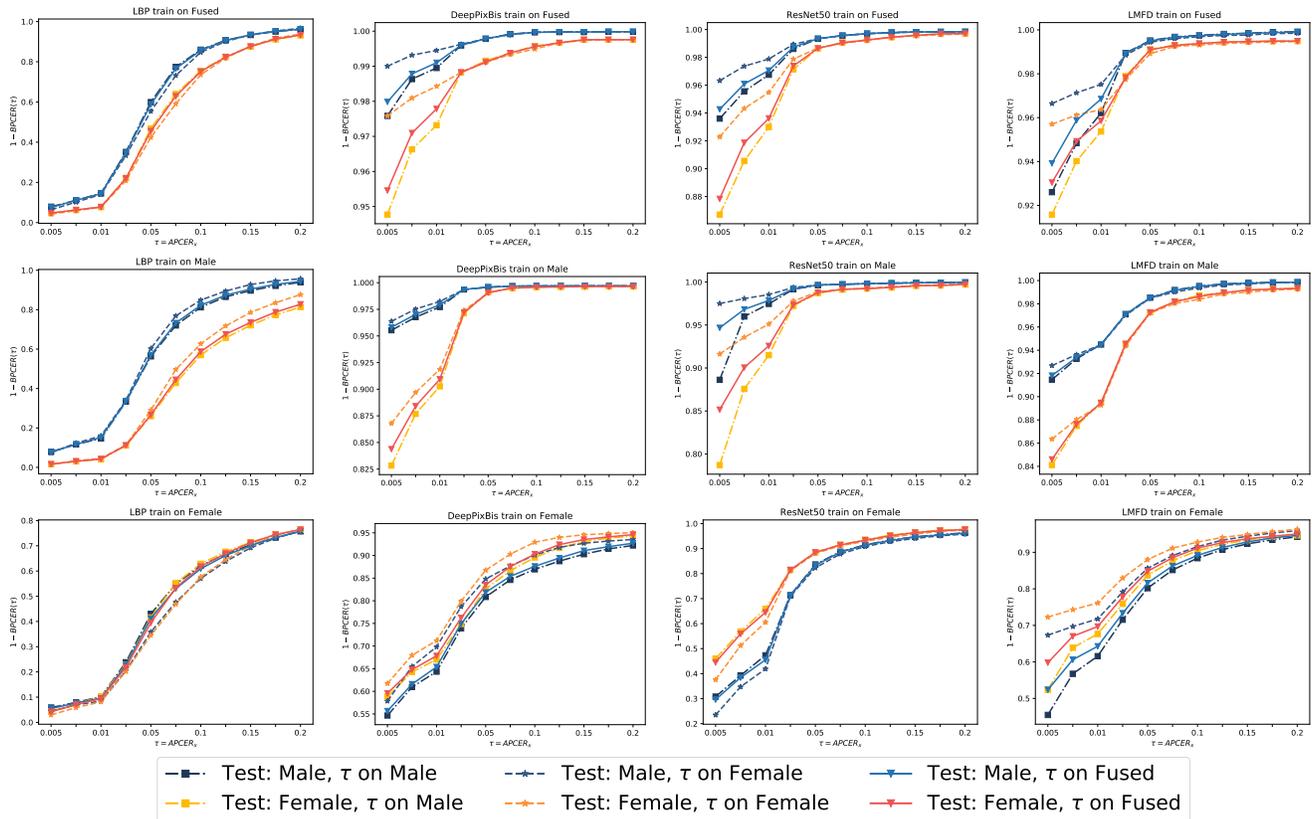

\centering
\foreach \model in {baseline,male,female}{
    \foreach \db in {LBP,DeepPixBis,ResNet50,LMFD}{
        \includegraphics[width=0.23\linewidth]{gender_PS_\db_\model.pdf}
    }
}
\includegraphics[width=0.8\linewidth]{gender_legend_male.pdf}
\caption{The results of FairSWAP models on gender group in terms of $BPCER\ @\ APCER_x$. While PAD performance differences between males and females still exist, the differences under different ODTAs are smaller, i.e. the curves within the blue and orange series are closer.}
\label{fig:bpcer_at_apcer_gender_ps}
\vspace{-5mm}
\end{figure*}

Table \ref{tab:eer_gender} presents the PAD performance in terms of EER values and Figure \ref{fig:bpcer_at_apcer_gender_ps} illustrates the performance under different ODTAs. 
%As shown in Table \ref{tab:eer_gender}, three deep-learning-based PAD methods with FairSWAP achieve improved performance in 13 out of 18 experimental cases on gender groups. 
{The results demonstrate that our FairSWAP solution enhances the fairness of deep-learning-based PAD methods in most cases (13 out of 18 experimental cases).}
However, LBP-MLP method did not benefit from FairSWAP. The possible reason is that the mixed illuminations in augmented images decrease the discrimination ability of color-channel-based LBP features.
As shown in Figure \ref{fig:bpcer_at_apcer_gender_ps}, the PAD performance differences between male and female groups still exist, but the differences under different ODTAs become smaller, which can be observed from the closer curves of similar colors (blue series or orange series). 

{The ABF curves in Figure \ref{fig:abf_gender} and FDR-AUC and ABF-AUC values presented in Table \ref{tab:fdr_abf_auc_gender} offer insights into the effectiveness of FairSWAP in boosting fairness. }
As shown in Figure \ref{fig:abf_gender}, the dotted ABF curves by FairSWAP are generally higher than the solid curves by baseline models, especially when models trained on the male group (blue curves). 
% Out of 72 cases, 44 cases show higher ABFs when considering ABFs at all plotted decision thresholds. 
Furthermore, {although LBP-MLP with FairSWAP does not obtain improved PAD performance, it exhibit higher fairness as reflected by higher FDR values} in Table \ref{tab:eer_gender}. 
FairSWAP consistently yields higher average ABF values than {the} baselines, {highlighting its potential benefits in enhancing both PAD performance and fairness.}

%=====================================================================
%=====================================================================

\subsection{Fairness enhancement over occlusion (Protocol-2)}
\label{ssec:ps_occlusion_results}
\begin{figure*}[htbp!]
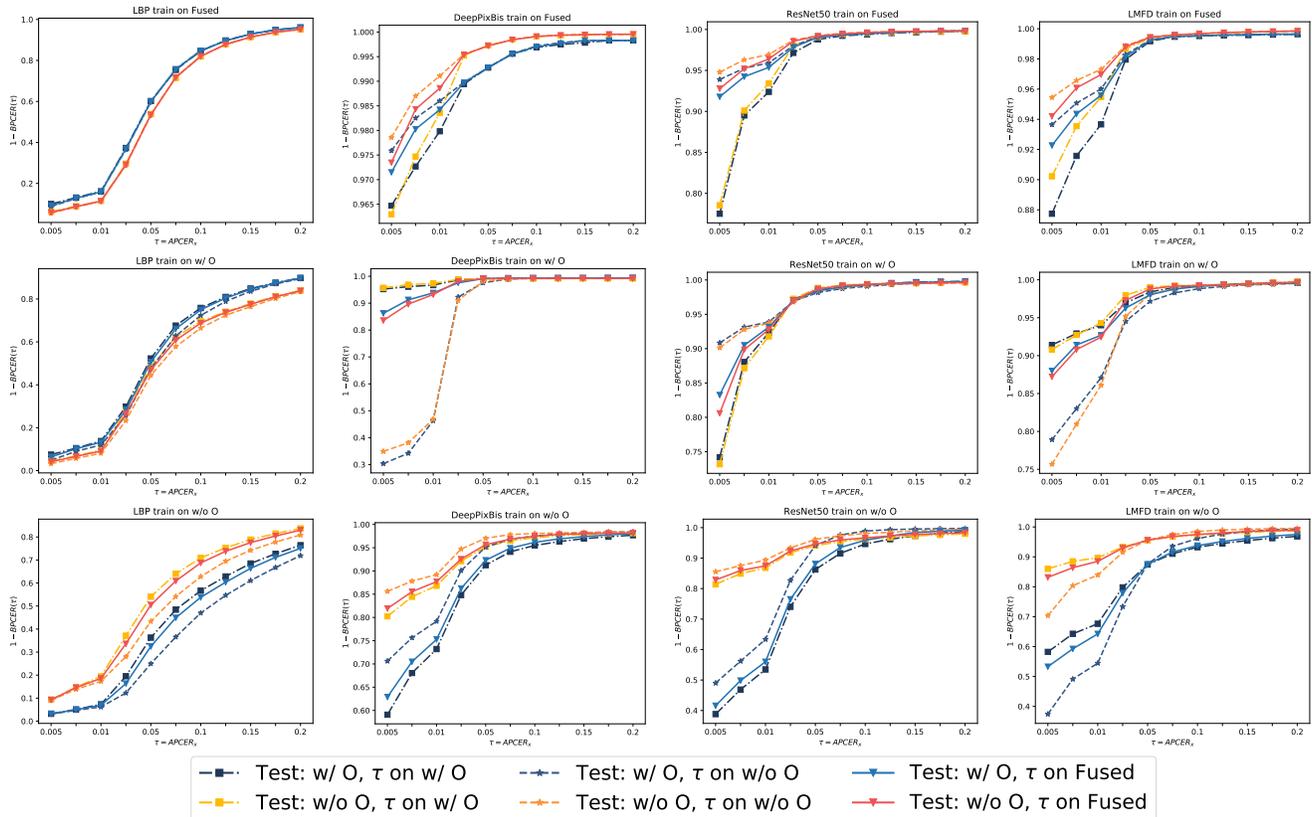

\centering
\foreach \model in {baseline,occlusion,noocclusion}{
    \foreach \db in {LBP,DeepPixBis,ResNet50,LMFD}{
        \includegraphics[width=0.23\linewidth]{occ_PS_\db_\model.pdf}
    }
}
\includegraphics[width=0.8\linewidth]{occ_legend_occlusion.pdf}
\caption{The results of FairSWAP on occlusion group in terms of $BPCER\ @\ APCER_x$. With the help of FairSWAP, the difference in PAD performance under different ODTAs is reduced, i.e. the curves are closer within the blue and orange series.}
\label{fig:bpcer_at_apcer_occlusion_ps}
\vspace{-5mm}
\end{figure*}

From Table \ref{tab:eer_occlusion} and Figure \ref{fig:bpcer_at_apcer_occlusion_ps}, we conclude that: 1) LBP-MLP method does not benefit much from FairSWAP compared to the deep-learning-based models. 2) FairSWAP significantly outperforms baseline models in seven out of eight cases when models trained on the non-occlusion group. 3) {FairSWAP decreases} the differential performance and outcome of PAD under different ODTAs. 
As shown in Figure \ref{fig:abf_occlusion}, the dotted ABF curves are smoother than {the} solid curves when LBP-MLP, DeepPixBis,and ReseNet50 are trained on the occlusion group (blue curves).
For models trained on the non-occlusion group, {the} ABF curves of three deep-learning-based models with FairSWAP (dotted green lines) are higher than {the} baselines (solid green lines). {These visual observations are consistent with the results} in Table \ref{tab:fdr_abf_auc_occlusion}. Deep-learning-based methods with FairSWAP result in higher average FDR-AUC and ABF-AUC values in all cases. In general, FairSWAP enhances the PAD performance and fairness in most cases, particularly for deep-learning-based PAD solutions.

%=====================================================================
%=====================================================================

\subsection{Fairness enhancement over other attributes (Protocol-3)}
\label{ssec:ps_other_attributes_results}

Table \ref{tab:eer_other} compares the PAD performance of baseline models and models incorporating FairWAP in terms of EER. {The results show that FairSWAP improves the PAD performance on most attribute groups, outperforming baseline models when considering the average performance.}
{Moreover, Table \ref{tab:fdr_abf_auc_attribute} demonstrates that FairSWAP enhances the fairness of PAD models,} as indicated by higher FDR-AUC values compared to the baselines.
However, LBP-MLP with FairSWAP obtains a lower average ABF-AUC values than without FairSWAP, mainly due to the behavior with the Makeup attribute. Deep-learning-based models particularly benefit from FairSWAP, {as evidenced by their higher ABF-AUC values.}
Overall, our FairSWAP approach enhances the PAD performance and fairness on most attribute groups, especially for deep-learning-based PAD solutions.

\section{Conclusion}
\label{sec:conclusion}
% adding some findings of the observation
% 1. distribution of training data cause bias
% 2. operational conditions also affect results
% 3. differential performance on 

% future work:
% 1. well-distributed dataset with more annotations
% 2. standard fairness definition and metrics
% 3. more fairness aspects, such as model training bias
This work provided a comprehensive analysis of fairness in face PAD. In Section \ref{sec:caad_pad}, we first presented the CAAD-PAD dataset by combining six publicly available face PAD datasets. To enable the fairness analysis, we provided seven human-annotated attribute labels, covering demographic and non-demographic attributes.
Then, we analyzed the fairness of face PAD and its relation to the nature of the training data and the ODTA by employing four face PAD methods. We further introduced a novel metric, ABF, to jointly represent both, the PAD fairness and the absolute PAD performance. Extensive experimental results in Section \ref{sec:fairness_assessment_baselines} pointed out that the training data and ODTA are triggers of unfairness in face PAD. Consequently, we propose a simple yet effective solution, FairSWAP, acting as a data augmentation technique, to enhance the fairness of face PAD. FairSWAP aims to disorder the attribute information and guide models to mine discriminative attack features instead of identity or appearance features. Extensive experiments on CAAD-PAD in Section \ref{sec:bias_mitigation_PS} demonstrated that FairSWAP boosts PAD performance and fairness on gender, occlusion, and other attribute groups in most experimental setups.

\noindent \textbf{Acknowledgments:}
This research work has been funded by the German Federal Ministry of Education and Research and the Hessen State Ministry for Higher Education, Research and the Arts within their joint support of the National Research Center for Applied Cybersecurity ATHENE.

%\section*{References}

%\section*{References}

%{\setstretch{1.3}\small\bibliography{main}}

{\bibliography{main}}

\end{document}